\documentclass[pdflatex,sn-mathphys-num]{sn-jnl}

\usepackage{graphicx}%
\usepackage{multirow}%
\usepackage{amsmath,amssymb,amsfonts}%
\usepackage{amsthm}%
\usepackage{mathrsfs}%
\usepackage[title]{appendix}%
\usepackage{xcolor}%
\usepackage{textcomp}%
\usepackage{manyfoot}%
\usepackage{booktabs}%
\usepackage{algorithm}%
\usepackage{algorithmicx}%
\usepackage{algpseudocode}%
\usepackage{listings}%
\usepackage{caption}

\theoremstyle{thmstyleone}%
\theoremstyle{thmstyletwo}%

\theoremstyle{thmstylethree}%

\title{Automated Global Analysis of Experimental Dynamics through Low-Dimensional Linear Embeddings}
\author{Samuel A. Moore$^{1}$}
\author{Brian P. Mann$^{1}$}
\author{Boyuan Chen$^{1,2,3}$}
\affil{$^{1}$Department of Mechanical Engineering and Materials Science, Duke University, Durham, North Carolina}
\affil{$^{2}$Department of Electrical and Computer Engineering, Duke University, Durham, North Carolina}
\affil{$^{3}$Department of Computer Science, Duke University, Durham, North Carolina}
\begin{document}
\maketitle

\begin{center}
    \color{blue}{\url{http://generalroboticslab.com/AutomatedGlobalAnalysis}}
\end{center}
\vspace{10mm}
\begin{abstract}
    \textbf{Dynamical systems theory has long provided a foundation for understanding evolving phenomena across scientific domains. Yet, the application of this theory to complex real-world systems remains challenging due to issues in mathematical modeling, nonlinearity, and high dimensionality. In this work, we introduce a data-driven computational framework to derive low-dimensional linear models for nonlinear dynamical systems directly from raw experimental data. This framework enables global stability analysis through interpretable linear models that capture the underlying system structure. Our approach employs time-delay embedding, physics-informed deep autoencoders, and annealing-based regularization to identify novel low-dimensional coordinate representations, unlocking insights across a variety of simulated and previously unstudied experimental dynamical systems. These new coordinate representations enable accurate long-horizon predictions and automatic identification of intricate invariant sets while providing empirical stability guarantees. Our method offers a promising pathway to analyze complex dynamical behaviors across fields such as physics, climate science, and engineering, with broad implications for understanding nonlinear systems in the real world.}
\end{abstract}

\newpage
\section*{Introduction}
Since Isaac Newton published \textit{Philosophiae Naturalis Principia Mathematica} in 1687 and society entered the Age of Enlightenment, the study of dynamics has shaped our understanding of the natural world. Initially focused on the forces between bodies, this paradigm has evolved into the broader framework of Dynamical Systems Theory. This theory now extends beyond the motion of physical bodies to encompass the study of time-varying state variables in diverse fields, including mechanical and electrical engineering, climate science, neuroscience, physiology, and ecology.

Analyzing real-world dynamical systems remains challenging due to difficulties in modeling, nonlinearity, and high dimensionality. For instance, although researchers can often measure systems, they frequently struggle to identify the underlying dynamics. Even when models are accurate, nonlinearity can restrict analysis to local regions of the state space, making it harder to understand the system's overall behavior \cite{strogatz2018nonlinear, khalil2002control, jordan2007nonlinear}. Additionally, high dimensionality—characterized by a large number of states in a system—can hinder meaningful interpretation and analysis.
\captionsetup[figure]{labelformat=empty}
\begin{figure}
    \centering
    \includegraphics[width=\textwidth]{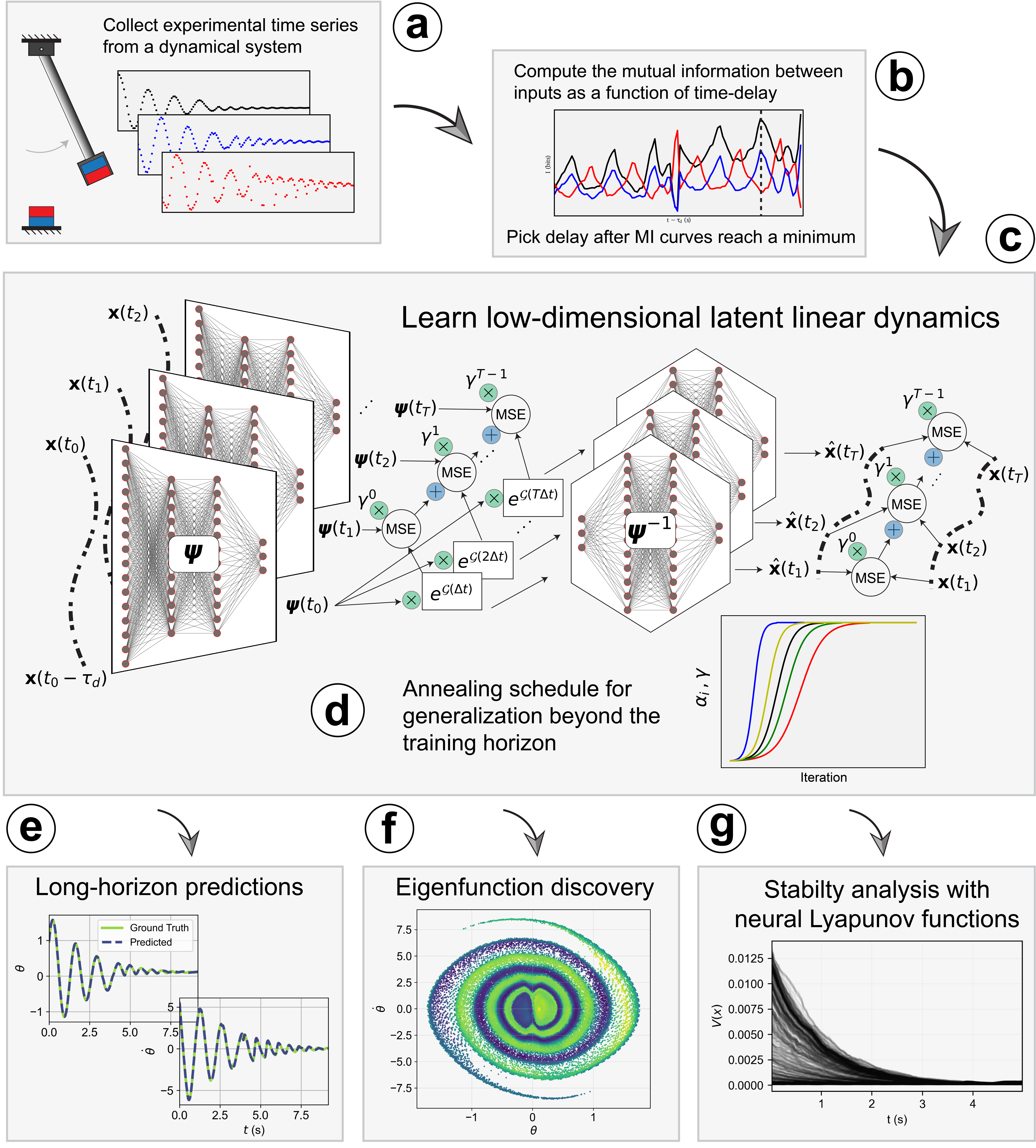}
    \caption{\textbf{Fig. 1 $\vert$ Automated global analysis of experimental dynamics.} An overview of our framework to automate the global analysis of experimental dynamical systems by learning low-dimensional latent linear embeddings. \textbf{a}, Collect time-series from a dynamical system. \textbf{b}, Choose the model input dimension by selecting an appropriate time-delay using the mutual-information between trajectories in the system. \textbf{c}, Train a deep autoencoder network that constrains the latent space to behave like a linear dynamical system. \textbf{d}, During training, annealing the coefficient of the loss function and the training prediction horizon to ensure model generalization. \textbf{e}, Long-horizon predictions. \textbf{f}, Interpretable eigenfunction discovery. \textbf{g}, Stability analysis with learned Lyapunov functions.}
    \label{fig: LLLD framework}
\end{figure}
In 1931, Bernard Koopman showed that a suitable change of coordinates could globally linearize nonlinear dynamics, offering a simple mathematical structure, albeit in an infinite-dimensional space \cite{koopman1931hamiltonian}. Unlike general dynamical systems, linear systems allow for straightforward global analysis through spectral decomposition and provide numerous control options \cite{hespanha2018linear, brunton2021modern, borrelli2017predictive}. Inspired by Koopman's work, recent research demonstrates that low-dimensional exact linearizations are achievable through eigenfunctions \cite{brunton2016koopman, brunton2021modern}. Beyond being low-dimensional and linear, eigenfunction coordinates can uncover hidden properties of the system, such as attractor structures and Lyapunov functions, which are not easily found in other coordinate representations and traditional techniques in dynamical systems. This makes eigenfunction coordinates a powerful technique for automated global analysis of nonlinear dynamics \cite{mauroy2016global, mauroy2013isostables, deka2022koopman, mezic2021koopman}.

In data science, researchers consider dimensionality one of the most important characteristics of data. They view high-dimensional data as an expression of a low-dimensional underlying manifold with an intrinsic dimension. As a result, researchers propose various techniques for dimensionality reduction and estimation \cite{vidal2005generalized, roweis2000nonlinear, rumelhart1986learning, levina2004maximum, camastra2016intrinsic}. Although the significance of dimensionality in data science is widely recognized, researchers still understand little about its role in Koopman-inspired models for dynamical systems. Moreover, whether low-dimensional eigenfunction representations exist for most systems remains unclear.

Researchers propose numerous methods for finding approximate linear models of nonlinear dynamics, starting with the now ubiquitous Dynamic Mode Decomposition (DMD) and extended Dynamic Mode Decomposition (eDMD) \cite{schmid2010dynamic, williams2015data}. Although these methods are generally straightforward to implement, DMD struggles with nonlinear dynamics, and eDMD often produces representations that are much higher dimensional than the original state space, leading to the curse of dimensionality \cite{schmid2010dynamic, williams2015data}.

Deep learning has become an effective tool for nonlinear dimensionality reduction, learning structured latent representations of data, and modeling dynamical systems \cite{tschannen2018recent, hafner2019learning, cranmer2020lagrangian, chen2018neural, ansuini2019intrinsic}. A prime example is the use of deep convolutional autoencoders to discover the intrinsic dimension of dynamical systems directly from high-dimensional video observations \cite{chen2022automated, huang2024automated}. Similarly to this line of research, our work seeks an alternative set of variables to describe dynamical systems beyond the space in which they are measured.

Researchers have used deep learning to find linear embeddings for nonlinear dynamics \cite{lusch2018deep, takeishi2017learning, han2020deep, yeung2019learning, shi2022deep, morton2019deep, liu2022physics}. However, much like DMD-based methods, these approaches still struggle to produce low-dimensional linear models for even simple nonlinear systems. For example, researchers used deep learning to find a 100-dimensional (100D) linear embedding space for the Duffing equation, a 2D bi-stable nonlinear system \cite{takeishi2017learning}. Similarly, they used 21D and 1000D embedding spaces with eDMD for the same system \cite{williams2015data, peitz2020data}. Another widely studied benchmark nonlinear system, the Van der Pol Oscillator, was represented as a 100D system \cite{iacob2021deep}, a 20D system \cite{deka2022koopman, korda2020optimal}, and a 28D linear system \cite{kamb2020time}, among others.

While the examples above suggest that high-dimensional linear embeddings can model nonlinear dynamics, the few analytical examples of Koopman eigenfunctions demonstrate the potential of capturing the dynamics with only a few key variables \cite{brunton2021modern}. Moreover, modeling systems with low-dimensional representations in higher-dimensional spaces increases the likelihood of redundancy, spurious modes, and overfitting. However, current methods for discovering eigenfunctions either lack representational capacity or fail to address dimensionality altogether \cite{korda2020optimal, kaiser2021data, folkestad2020extended, deka2022koopman, haseli2021learning}.

We demonstrate that low-dimensional linearization is possible for a wide class of nonlinear systems using our novel data-driven machine learning approach. Specifically, we show that 3D and 6D representations are sufficient to accurately model the Van der Pol and Duffing oscillators, respectively. Beyond these prototypical systems, we extend our study to numerous previously unstudied, experimental and real-world nonlinear dynamical systems by discovering their low-dimensional linear representations. Our results demonstrate that this approach improves generalization, enables long-horizon predictions, reduces the occurrence of false modes, and facilitates interpretability, empirical stability analysis, and the discovery of intricate invariant sets.
\captionsetup[figure]{labelformat=empty}
\begin{figure}[h!]
    \centering
    \includegraphics[width=\textwidth]{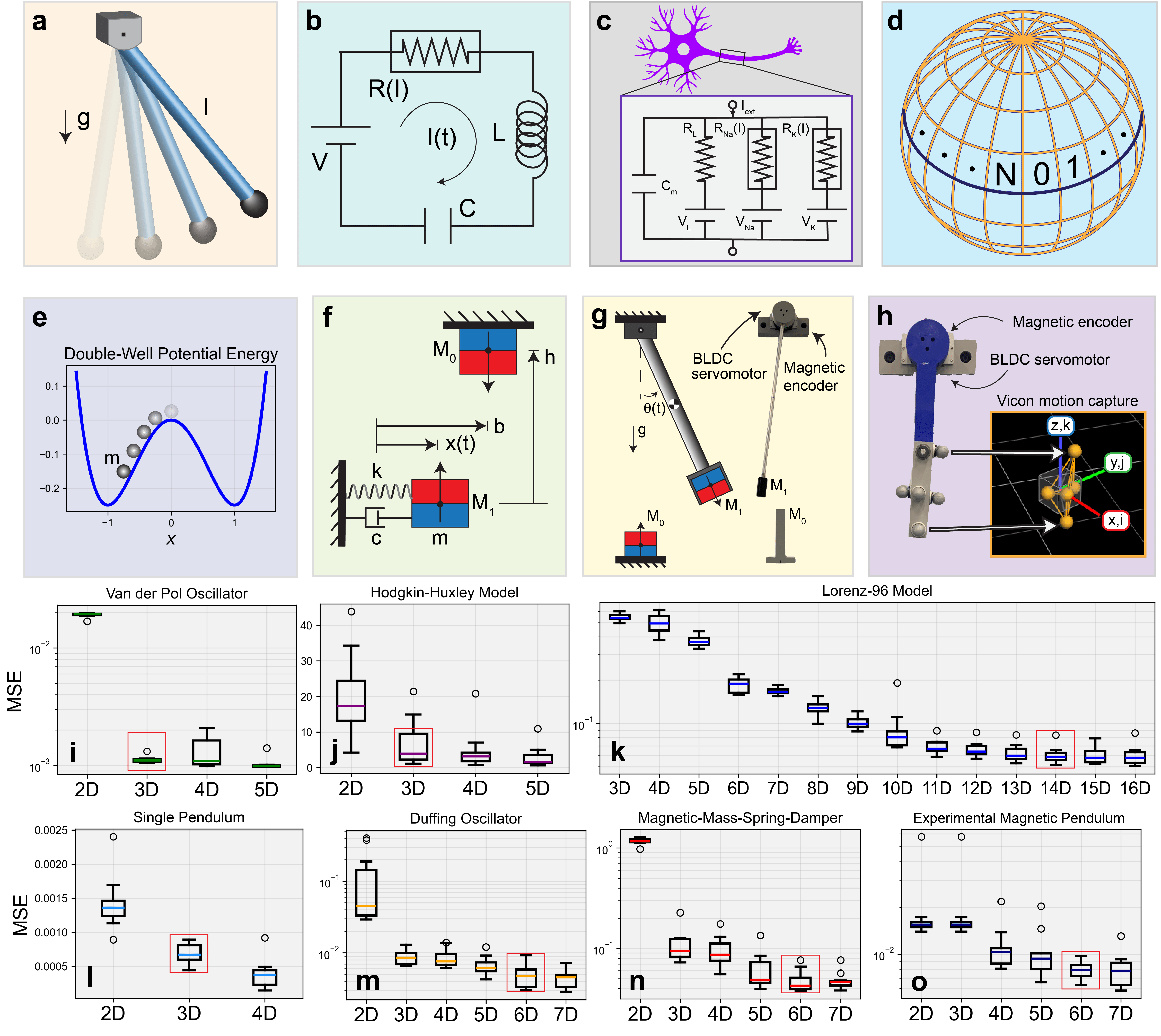}
    \caption{\textbf{Fig. 2 $\vert$ Datasets and prediction error.} Diagrams detailing the studied dynamical systems and the prediction error as a function of latent dimension. \textbf{a}, A single pendulum model. \textbf{b}, A circuit with nonlinear resistance known as the Van der Pol oscillator. \textbf{c}, A model for how action potential in neurons are initiated and propagated called the Hodgkin-Huxley Model. \textbf{d}, A model which was devised to study weather predictability known as the Lorenz-96 system. \textbf{e}, A particle mass situated in a double well potential called the Duffing oscillator. \textbf{f}, A mass-spring-damper system with two repelling magnets. \textbf{g}, An experimental magnetic pendulum. \textbf{h}, An experimental double pendulum. \textbf{i-o}, Box and whisker plots showing the mean squared prediction error across embedding dimensions for some of the studied systems. Using our learning approach, the prediction error plateaus with a relatively low-dimensional state space. The red box indicates the latent dimension that the system was modelled in.}
    \label{fig: studied systems}
\end{figure}
We achieve these results with a multistep procedure and a deep autoencoder network with physics-informed optimization to structure the latent space in accordance with Koopman Operator Theory (Fig. \ref{fig: LLLD framework}). This approach works directly from experimental data (Fig. \ref{fig: LLLD framework}a). We use time-delay embedding as a central feature to enhance prediction performance and implement a method to select the length of time-delay before training using mutual information (Fig. \ref{fig: LLLD framework}b). While fitting the networks to learn long-horizon predictions (Fig. \ref{fig: LLLD framework}c) we apply regularization through hyperparameter annealing (Fig. \ref{fig: LLLD framework}d). Once the model is trained, we perform long-horizon predictions (Fig. \ref{fig: LLLD framework}e), discover eigenfunctions (Fig. \ref{fig: LLLD framework}f), and conduct empirical global stability analysis (Fig. \ref{fig: LLLD framework}g). With this framework, we discover entirely new representations for electrical systems, neural circuits, magnetic pendulums, and atmospheric processes. Consequently, our work can facilitate future discoveries in fields such as physics, robotics, biology, and neuroscience, where the choice of coordinate representation is non-trivial, the behaviors are poorly understood, and experimental data is abundant.

\section*{Results}
\subsection*{Studied Dynamical Systems}
We created nine datasets across multiple scientific fields using trajectories from simulated and experimental nonlinear dynamical systems (Fig. \ref{fig: studied systems}, \hyperref[sec:supplementary materials a]{Supp. Section A}). The studied systems exhibit a range of behaviors, from relatively simple to highly complex. The single pendulum was the simplest system studied, with only two state variables describing its angular position and velocity (Fig. \ref{fig: studied systems}a, \hyperref[sec:supplementary materials a]{Supp. Section A}). Its simplicity is further highlighted by the presence of a single fixed-point attractor. In contrast, the more complex Van der Pol oscillator dataset features a limit-cycle attractor and was introduced by Dutch electrical engineer Balthasar van der Pol while studying nonlinear circuits in vacuum tubes (Fig. \ref{fig: studied systems}b, \hyperref[sec:supplementary materials a]{Supp. Section A}) \cite{van1926lxxxviii,  van1927frequency}. The Hodgkin-Huxley model, introduced by Alan Hodgkin and Andrew Huxley in 1952 to describe the excitation mechanisms of neurons, features four state variables, strong nonlinearity, and self-sustained oscillations (Fig. \ref{fig: studied systems}c, \hyperref[sec:supplementary materials a]{Supp. Section A}). The mechanisms uncovered by this model were foundational in advancing our understanding of neural excitation, earning Hodgkin and Huxley the Nobel Prize in Physiology or Medicine in 1963 \cite{hodgkin1952quantitative}. The Lorenz-96 system, developed by Edward Lorenz, the founder of chaos theory, served as the next dataset (Fig. \ref{fig: studied systems}d, \hyperref[sec:supplementary materials a]{Supp. Section A}). Originally conceived to explore weather predictability, this model was characterized by its high-dimensional state space and its ability to exhibit both periodic and chaotic solutions \cite{lorenz1996predictability}. 

We also constructed another four datasets from dynamical systems characterized by multistability, in contrast to the previously discussed models. The initial dataset was derived from the Duffing oscillator, introduced by George Duffing in 1918 to analyze mechanical vibrations \cite{duffing1918erzwungene}. Often depicted as a particle mass situated in a double-well potential energy landscape, this system displays more complex stability behavior compared to the earlier datasets (Fig. \ref{fig: studied systems}e, \hyperref[sec:supplementary materials a]{Supp. Section A}). The subsequent dataset came from a model involving two interacting magnetic dipoles: one fixed, and the other free to oscillate while attached to a spring and damper (Fig. \ref{fig: studied systems}f, \hyperref[sec:supplementary materials a]{Supp. Section A}). This model exhibited pronounced nonlinear behavior due to magnetic repulsion compared to the Duffing oscillator and featured particularly asymmetric basins of attraction \cite{wang2022model}. We also examined the dynamics of a system with multiple nested limit cycles, as opposed to fixed points (\hyperref[sec:supplementary materials a]{Supp. Section A}). Lastly, we constructed two experimental pendulum datasets, the first was derived from a single pendulum with magnetically induced multi-stability and the second was from a double pendulum that exhibited chaotic behavior (Fig. \ref{fig: studied systems}g,h, \hyperref[sec:supplementary materials a]{Supp. Section A}). 

\captionsetup[figure]{labelformat=empty}
\begin{figure}[H]
    \centering
    \includegraphics[width=\textwidth]{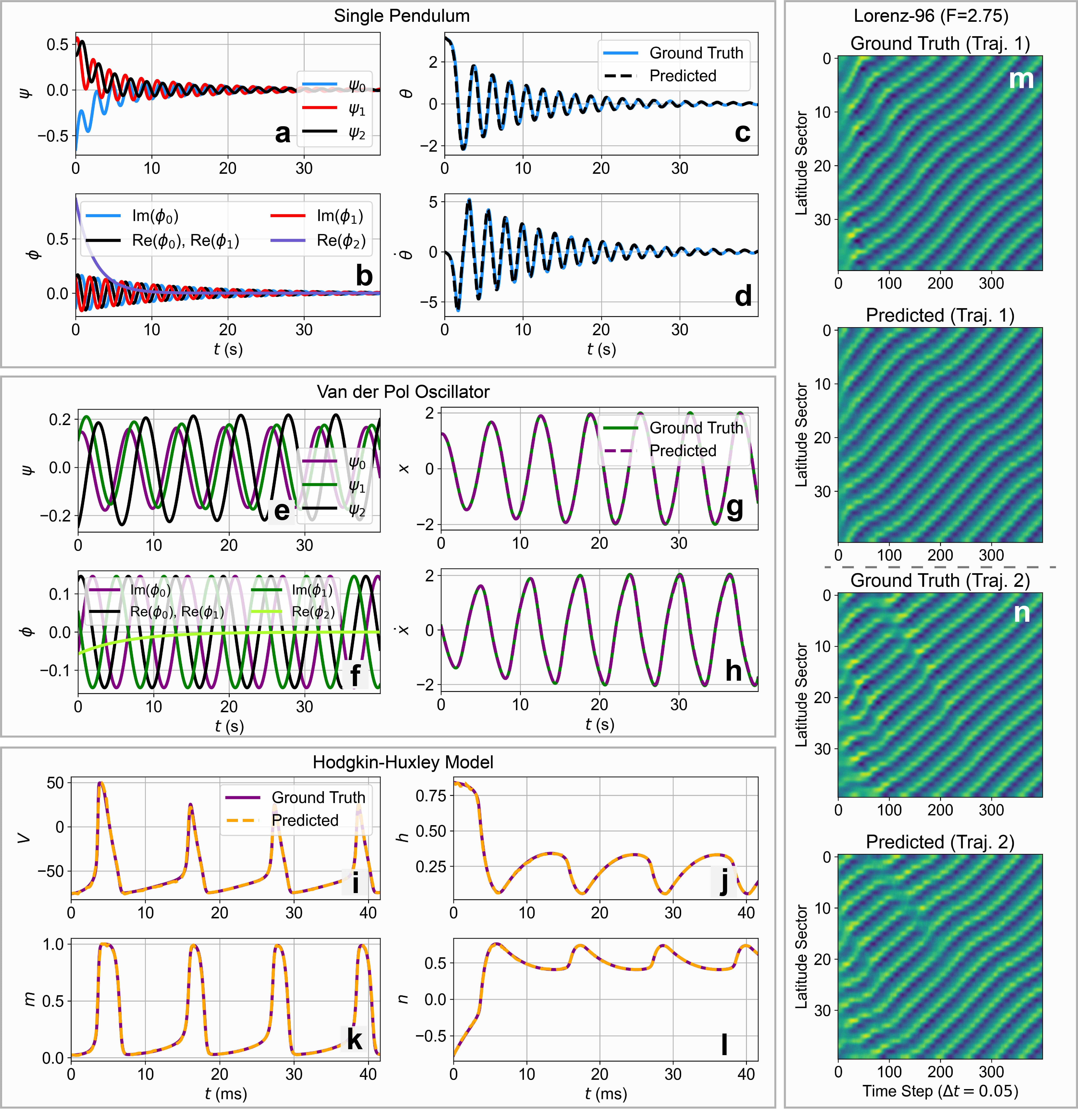}
    \caption{\textbf{Fig. 3 $\vert$ Long-horizon predictions.} Predicted trajectories from low-dimensional linear embeddings of nonlinear dynamics. \textbf{a}, Predicted trajectories in latent space for the single pendulum modeled as a 3D linear system. \textbf{b}, The same latent space trajectories for the pendulum, decomposed into separate modes. \textbf{c} and \textbf{d}, Ground truth and predicted trajectories for angular position and velocity after decoding into state space. \textbf{e}, Predicted latent states for the Van der Pol oscillator as a 3D linear system. \textbf{f}, The same latent space trajectories for the Van der Pol oscillator, decomposed into separate modes. \textbf{g} and \textbf{h}, Ground truth and predicted trajectories for the state variables of the Van der Pol oscillator after decoding. \textbf{i-l}, Ground truth and predicted trajectories in state space for the Hodgkin-Huxley model as a 3D linear system. \textbf{m} and \textbf{n}, Future predicted and ground truth states for the periodic Lorenz-96 model with 40 latitude sectors, represented as a 14D linear system.}
    \label{fig: predictions 1}
\end{figure}
\subsection*{Long-Horizon Predictions with Low-Dimensional Linear Dynamics}
We first evaluated the ability of our approach to make accurate long-horizon predictions, which is the first step towards trusting the model for system analysis and knowledge discovery. We developed a set of algorithmic components to achieve accurate long-horizon predictions including time-delay, loss functions, curriculum learning, and regularization techniques. During training and inference, we conducted prediction rollouts in the latent space, $\boldsymbol{\psi}$, after encoding an initial time-delayed state. The length of time-delay had a large impact on future prediction error (\hyperref[fig:extended data fig 1]{Extended Data Fig. 1}a,b). However, it remains unclear on how to select the time-delay parameter in existing data-driven methods. Therefore, a method to select the time-delay in our latent linear models was developed based mutual-information (Fig. \ref{fig: LLLD framework}b, \hyperref[fig:extended data fig 1]{Extended Data Fig. 1}b,c,  \hyperref[sec:methods]{Methods}), inspired by traditional dynamical systems research in time-delay embedding \cite{fraser1986independent, fraser1989information, abarbanel1994predicting, abarbanel2012analysis}.

We performed model rollouts using the analytical solution to our learned linear model for the dynamics  (Fig. \ref{fig: LLLD framework}c, \hyperref[sec:methods]{Methods}). To ensure accurate predictions, we supervised these rollouts using ground truth future embeddings over $T$ time steps, progressively scaling down each prediction based on its temporal distance from the present using a discount factor $\gamma$ (Fig. \ref{fig: LLLD framework}c, \hyperref[sec:methods]{Methods}).  After generating the latent space predictions, we re-projected them into state space using $\boldsymbol{\psi}^{-1}$, and during training, we supervised these predictions with ground truth future states (Fig. \ref{fig: LLLD framework}c, \hyperref[sec:methods]{Methods}). 

Obtaining predictions that generalized beyond the training horizon was essential for learning accurate dynamics in a low-dimensional latent space. We achieved this prediction generalization capability through systematic annealing of the discount factor, which we implemented as a form of curriculum learning for the prediction horizon (Fig. \ref{fig: LLLD framework}d, \hyperref[sec:methods]{Methods}). By training with a variable horizon, we demonstrated nearly two orders of magnitude improvement in long-horizon predictions compared to the fixed-horizon approaches used in previous work (\hyperref[fig:extended data fig 1]{Extended Data Fig. 1}). To select the embedding dimension for each system, we examined the validation prediction error of our trained models across dimensions (Fig. \ref{fig: studied systems}i-o). As part of our regularization strategy and to keep the model parsimonious, we chose the smallest embedding dimension that did not significantly degrade performance.

We represented the single pendulum as a 3D linear system. Although the prediction error decreased slightly with the addition of another latent state, the error was already exceptionally small in 3D. We visualized example predictions in latent space for a trajectory starting at the upright position of the pendulum (Fig. \ref{fig: predictions 1}a). Leveraging the linear evolution of the latent states we decomposed the predicted latent states into separate dynamic modes, $\phi$, of oscillation, growth, and decay with Koopman Mode Decomposition (Fig. \ref{fig: predictions 1}b, \hyperref[sec:methods]{Methods}). The predictions in state space remained accurate across all future time-steps and successfully captured the frequency-shifting behavior in 3D (Fig. \ref{fig: predictions 1}c,d), as compared to the models trained in 2D (\hyperref[fig:extended data fig 2]{Extended Data Fig. 2}). (\hyperref[fig:extended data fig 2]{Extended Data Fig. 2}). Furthermore, we found that the 4D models introduced additional modes to the dynamics, which were likely to be false (\hyperref[fig:extended data fig 2]{Extended Data Fig. 2}). 

The prediction error for the Van der Pol oscillator plateaued after we lifted the latent space by a single dimension (Fig. \ref{fig: studied systems}i). Therefore, we modelled the Van der Pol oscillator as a 3D linear system, in stark contrast to the 100D, 28D, and 28D models in previous work \cite{iacob2021deep, deka2022koopman, korda2020optimal, kamb2020time}. As with the other systems, we integrated the dynamics ahead in time in the latent space (Fig. \ref{fig: predictions 1}e) and decomposed the predictions into separate dynamic modes (\ref{fig: predictions 1}f). When projected back into state space, our predictions closely matched the ground truth over an extended horizon, even capturing the transient behavior before settling on the limit-cycle attractor (Fig. \ref{fig: predictions 1}g,h). In contrast, the models trained in 2D failed to capture this transient behavior, while the 4D model exhibited an unstructured latent space, indicating overfitting (\hyperref[fig:extended data fig 2]{Extended Data Fig. 2}).
\begin{figure} 
  \centering 
  \includegraphics[width=\textwidth]{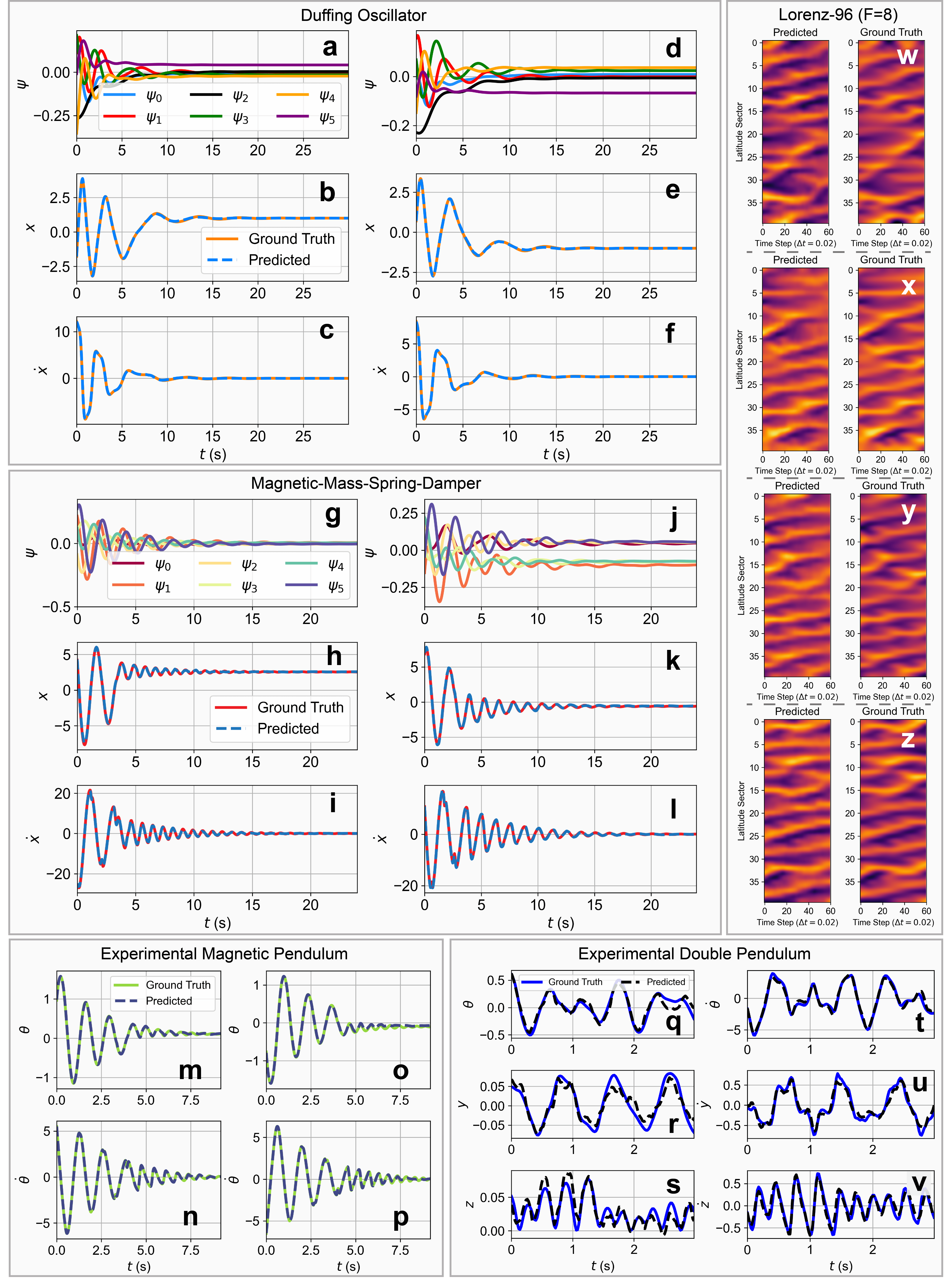}
  \caption{(Caption next page.)}
  \label{fig: predictions 2}
\end{figure}
\addtocounter{figure}{-1}
\captionsetup[figure]{labelformat=empty}
\begin{figure} 
    \caption{\textbf{Fig. 4$\vert$ Long-horizon predictions for multi-stable and chaotic systems.} \textbf{a}, The six learned latent states for the Duffing oscillator, forecasting over an extended horizon. \textbf{b} and \textbf{c}, Predicted and ground truth trajectories after decoding, including the correctly anticipated resting attractor. \textbf{d-f}, Forecasted latent variables and states for the Duffing oscillator that come to rest in the opposite potential well compared to the previous trajectory. \textbf{g-i}, Forecasted latent variables (in 6D) and states for the magnetic-mass-spring-damper system with asymmetric basins of attraction. \textbf{h} and \textbf{i}, Predicted and ground truth trajectories that come to rest in the smaller attractor. \textbf{g-i}, Predicted latent variables and states for the magnetic-mass-spring-damper system that come to rest in the larger attractor. \textbf{m-p}, Ground truth and predicted trajectories for each attractor of the experimental magnetic pendulum when modeled as a 6D linear system. \textbf{q-v}, Long-horizon predicted and ground truth trajectories for the measured states of the experimental double pendulum. \textbf{w-z}, Four forecasted and ground truth trajectories for the chaotic Lorenz-96 model with 40 states.
}
\end{figure}

Interestingly, we observed that the prediction error for the Hodgkin-Huxley model plateaued before reaching the original state dimension of four (Fig. \ref{fig: studied systems}j). This suggests redundancy in the original coordinate choice by Hodgkin and Huxley and confirms that our approach is capable of discovering novel coordinates for dimensionality reduction in addition to linearization. Like our model for the Van der Pol Oscillator, the model trained with the Hodgkin-Huxley dataset achieved accurate long-horizon predictions, including transient behavior (Fig. \ref{fig: predictions 1}i-l). We also reduced the dimensionality of the Lorenz-96 model, which exhibits limit-cycle behavior, from the original state dimension of 40 to 14 (Fig. \ref{fig: studied systems}k). Despite this substantial reduction, the predictions in the original state space aligned closely with the ground truth trajectories over a long horizon (Fig. \ref{fig: predictions 1}m,n).

We also achieved long-horizon prediction capabilities with low-dimensional linear models on systems with more than one attractor. We modelled the Duffing oscillator as a 6D linear system (Fig. \ref{fig: studied systems}m) compared to the 21D, 100D, and 1000D models used in previous work \cite{takeishi2017learning,  williams2015data, peitz2020data}. The predicted evolution of the six latent states, for a high energy initial condition, is shown in Fig. \ref{fig: predictions 2}a. After decoding the predictions, we accurately captured both inter-well and intra-well oscillations, as well as the correct resting attractor (Fig. \ref{fig: predictions 2}c,d). Predicting the intra-well oscillations was particularly challenging for the 4D model (\hyperref[fig:extended data fig 2]{Extended Data Fig. 2}). Additionally, we plotted a second trajectory starting from an initial condition close to the first in  Fig. \ref{fig: predictions 2}d. The model correctly inferred that the system comes to rest in the opposite attractor, even though the initial conditions were close in state space (Fig. \ref{fig: predictions 2}e,f). 

Similar to the Duffing oscillator, we observed the prediction error level-off at 6D in the simulated and experimental magnetic systems (Fig. \ref{fig: studied systems}n,o). Our approach successfully modelled the full scope nonlinear behavior in the magnetic-mass-spring-damper and magnetic pendulum systems. This included the impact-like velocity jumps as the magnets come in close proximity, the various frequencies of oscillation, and the multi-stability (Fig. \ref{fig: predictions 2}g-l). Lastly, chaotic systems, such as the experimental double pendulum, pose significant challenges for finite-dimensional Koopman models because of their spectral characteristics \cite{otto2021koopman}. Due to this, we observed little correlation between the prediction error and the embedding dimension (\hyperref[fig:extended data fig 2]{Extended Data Fig. 2}). While latent linear models may not perfectly match the spectral properties of chaotic systems, they still provided an interpretable and accurate framework for forecasting the future states in the experimental double pendulum and the chaotic Lorenz-96 model (Fig. \ref{fig: predictions 2}q-z).

\subsection*{Eigenfunction Discovery and Spectral Analysis}
Eigenfunctions play a crucial role in understanding physical phenomena across various fields, including quantum mechanics, vibration analysis, and thermal processes. Traditionally, eigenfunctions are derived from linear systems theory, which limits their direct application to nonlinear systems. However, because our approach results in a linear representation of nonlinear systems, we can directly apply modal decomposition to characterize the latent variables as complex-valued eigenfunctions, providing a detailed description of the system’s behavior. Specifically, the eigenvalues, eigenvectors, and eigenfunctions allowed us to analyze oscillation frequencies, phases, and growth and decay rates within each system (\hyperref[sec:methods]{Methods}). Additionally, we used the decaying modes of the system to construct neural Lyapunov functions (\hyperref[sec:methods]{Methods}) which characterized the attractive and equilibrium states of the system and provided stability guarantees. To ensure that the learned modes were physically realistic, we constructed a term to the loss function that penalized positive real-part eigenvalues, which would otherwise lead to unbounded exponential growth in the predicted trajectories (\hyperref[sec:methods]{Methods}).

\captionsetup[figure]{labelformat=empty}
\begin{figure}
    \centering
    \includegraphics[width=\textwidth]{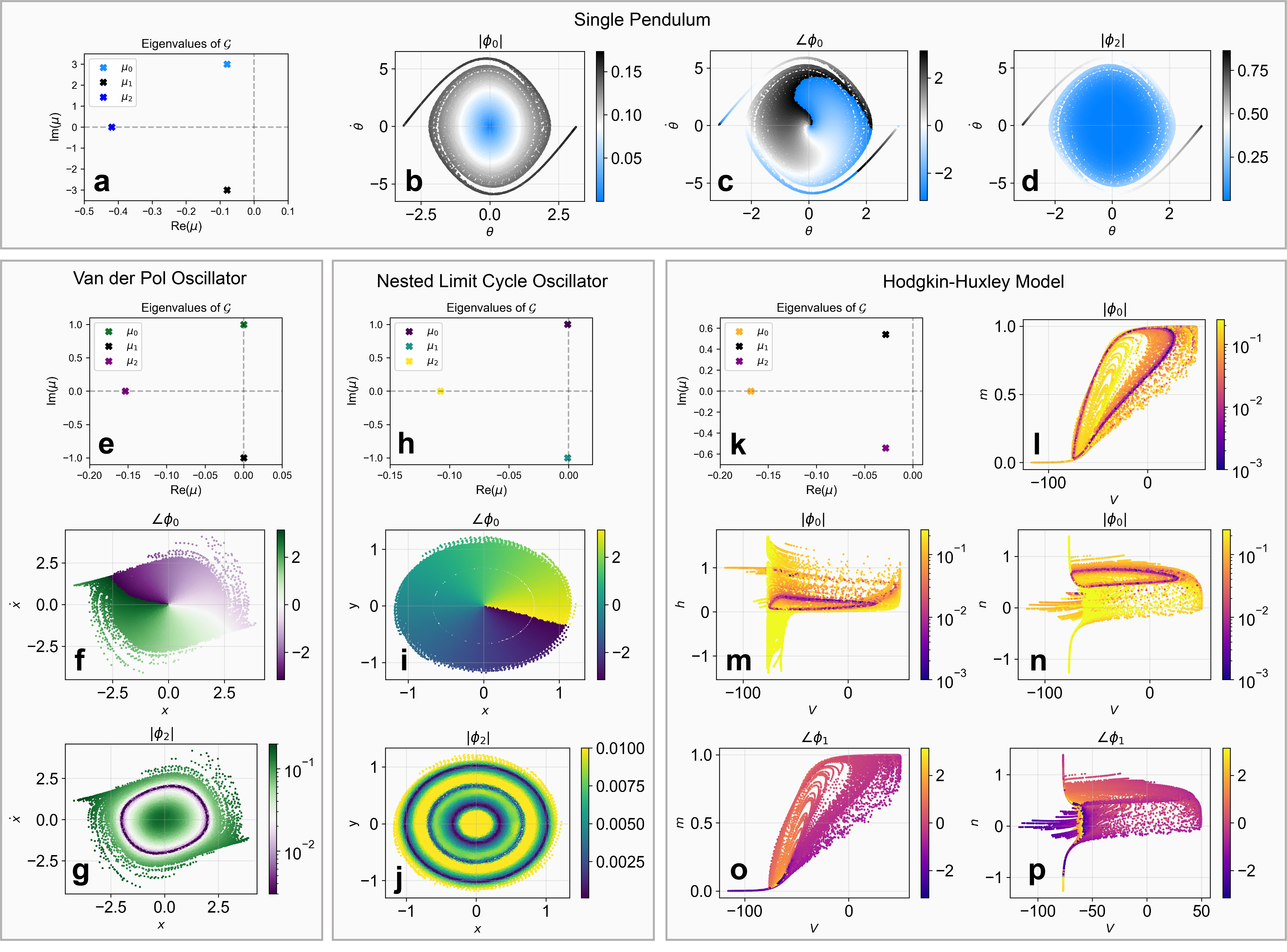}
        \caption{\textbf{Fig. 5 $\vert$ Learned eigenvalues and eigenfunctions.} \textbf{a}, The learned eigenvalues for the single pendulum model in a 3D embedding space. The complex part corresponds to the frequency of oscillation, and the real part corresponds to decay (or growth if positive). \textbf{b} and \textbf{c}, The magnitude and phase of one of the complex-valued, learned eigenfunctions as a function of the input states. \textbf{d}, The magnitude of the purely real, learned eigenfunction. \textbf{e}, The learned eigenvalues for the Van der Pol oscillator in a 3D embedding space. \textbf{f} and \textbf{g}, The phase of one of the purely complex-valued eigenfunctions and the magnitude of the purely real-valued eigenfunction. \textbf{h-j}, The learned eigenvalues and eigenfunctions for a model with three nested limit cycle attractors as a 3D linear system. \textbf{k-p}, The learned eigenvalues and eigenfunctions for the Hodgkin-Huxley model as a 3D linear system. Some of the depicted color maps are clamped for visual clarity.} 
    \label{fig: eigenfunction 1}
\end{figure}

We first examined the spectrum of the pendulum and identified a complex conjugate pair of eigenvalues, $\mu_{0, 1} = \alpha \pm \beta i$, and a purely real-valued eigenvalue $\mu_2$ (Fig. \ref{fig: eigenfunction 1}a). The real-part of the eigenvalues were negative, which corresponded to global exponential decay in the system. The imaginary part of the complex-conjugate pair indicated the natural frequencies in latent space (\hyperref[sec:methods]{Methods}). We explored the system's eigenfunctions as functions of the input states (Fig. \ref{fig: eigenfunction 1}b-d) and as trajectories (Fig. \ref{fig: predictions 1}b). When we examined the magnitude of the eigenfunction $\phi_0$, which corresponded to $\mu_0$, it revealed decay toward the pendulum’s downward position, $\theta = \dot{\theta} = 0$ (Fig. \ref{fig: eigenfunction 1}b). The minima in Koopman eigenfunctions indicates an equilibrium or attractor state in the dynamics (\hyperref[sec:methods]{Methods}), which we will further verify on the studied systems through empirical Lyapunov analysis in the next section. Notably, the equilibrium state in the pendulum was automatically revealed  as a feature of the learned latent space. Our model also provided further interpretation of the system through the phase of the oscillatory mode, which described how the periodic latent trajectory shifted as a function of the state variables (Fig. \ref{fig: eigenfunction 1}c). Lastly, we studied the magnitude of the final learned eigenfunction, $|\phi_2|$, which also identified decay toward the downward position, but at a faster rate (Fig. \ref{fig: eigenfunction 1}d).

We also performed spectral decomposition of the model for the Van der Pol oscillator, and  discovered a pair of purely oscillatory modes and a purely decaying mode (Fig. \ref{fig: eigenfunction 1}e). When we examined the phase of the oscillatory mode as a function of the inputs, it revealed an intricate symmetry (Fig. \ref{fig: eigenfunction 1}e). The magnitude of the purely real eigenvalue served as a neural Lyapunov function, allowing us to identify the limit-cycle attractor (Fig. \ref{fig: eigenfunction 1}g). Our approach automatically uncovered this attractor by modeling the Van der Pol oscillator as a 3D linear system, eliminating the need for a high-dimensional representation that would have required further refinement after learning \cite{deka2022koopman}. We observed a similar decomposition of modes with the more complex nested limit cycle and Hodgkin-Huxley models (Fig. \ref{fig: eigenfunction 1}h-p).

\begin{figure}
    \centering
    \includegraphics[width=\textwidth]{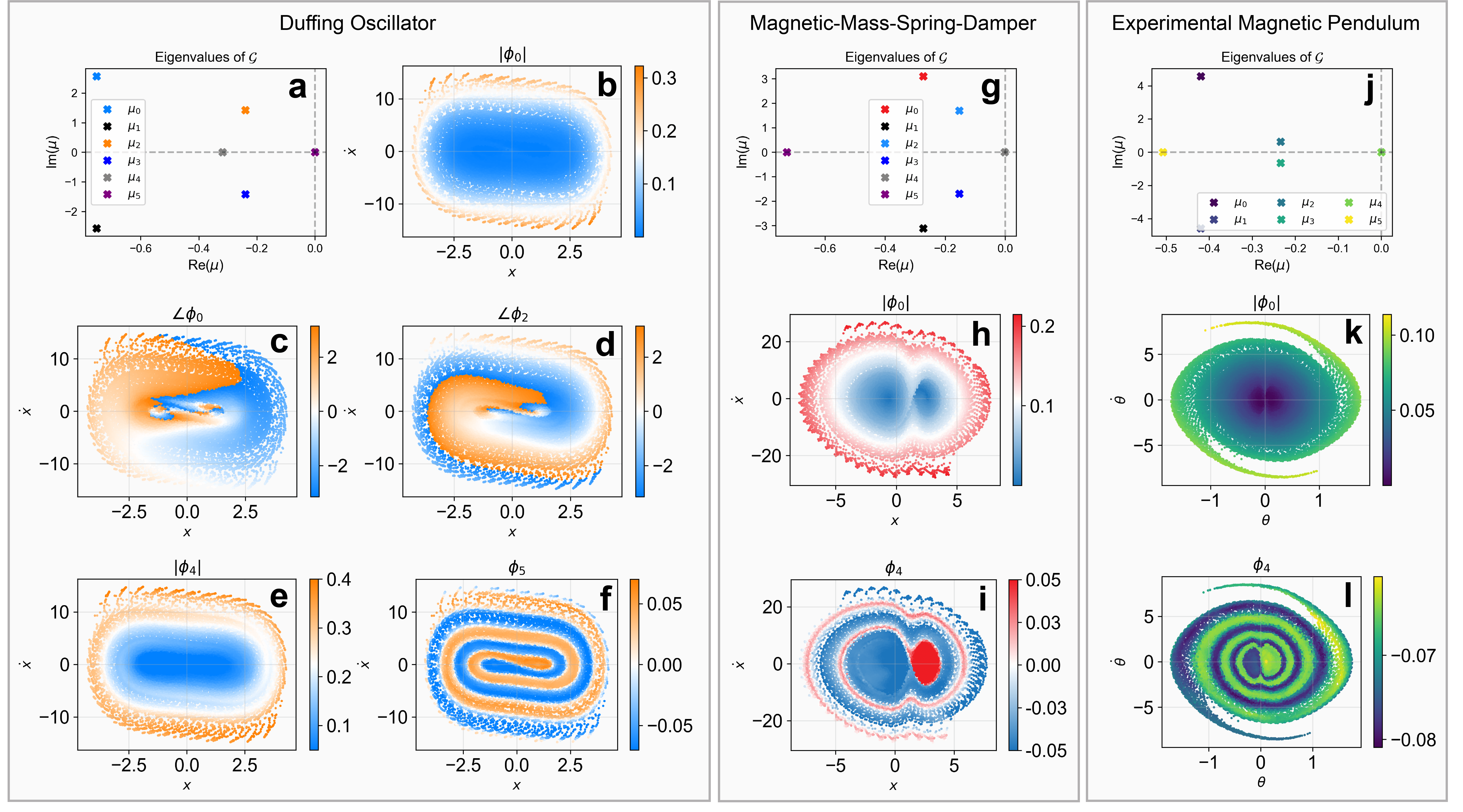}
        \caption{\textbf{Fig. 6 $\vert$ Learned eigenvalues and eigenfunctions for multi-stable systems.} \textbf{a}, The learned eigenvalues of the 6D latent linear model for the Duffing oscillator. \textbf{b-e}, The magnitude and phase of two complex-valued learned eigenfunctions for the Duffing oscillator. \textbf{f}, The learned eigenfunction for the system with the smallest magnitude eigenvalue. \textbf{g}, The six learned eigenvalues for the magnetic-mass-spring-damper system. \textbf{h}, The magnitude of one of the complex-valued eigenfunctions for the system. \textbf{i}, The learned eigenfunction for the magnetic-mass-spring-damper system with the smallest magnitude eigenvalue. \textbf{j-l}, The learned eigenvalues and two eigenfunctions from the experimental magnetic pendulum system.}
    \label{fig: eigenfunction 2}
\end{figure}

Most of the multi-stable systems required additional dimensions for accurate predictions, which lead to more modes of behavior compared to the mono-stable systems. We identified six eigenvalues for the Duffing oscillator, revealing four oscillatory and decaying modes, each characterized by two distinct frequencies and decay rates, along with a purely decaying mode and a static mode with negligible real or imaginary components (Fig. \ref{fig: eigenfunction 2}a). Similar to the previous systems, we used the decaying modes to construct neural Lyapunov functions and identified the equilibria of the system. By examining the phase of the various oscillation frequencies, we uncovered interesting patterns (Fig. \ref{fig: eigenfunction 2}b-e). Unlike the mono-stable systems, our learned model for the Duffing oscillator included a static mode that separated the state space into two parts corresponding to the basins of attraction (Fig. \ref{fig: eigenfunction 2}f). The basin of attraction for an equilibrium represents the set of initial conditions that lead to that equilibrium as $t \rightarrow \infty$. Previous methods have used eigenfunctions to identify the basins of attraction for the Duffing oscillator, but they were limited to small regions in state space and did not capture the inter-well oscillations that result in the spiral pattern seen in Fig. \ref{fig: eigenfunction 2}f \cite{takeishi2017learning, williams2015data}. Not only did our approach improve upon previous Koopman-based methods, but unlike brute-force techniques in traditional dynamical systems, we automatically discovered the basins of attraction as a continuous function of the original state variables.

Our learned model for the magnetic-mass-spring-damper had a similar structure to the Duffing oscillator with four oscillating and decaying modes, a decaying mode, and a static mode (Fig. \ref{fig: eigenfunction 2}g). We used the magnitude of the first mode, acting as a neural Lyapunov function, to reveal decay in the state space, indicating the presence of equilibrium states (Fig. \ref{fig: eigenfunction 2}h). By analyzing the static mode, we uncovered the vastly asymmetric basins of attraction (Fig. \ref{fig: eigenfunction 2}i), which validated that our approach is not limited to systems with simple nonlinearities like the Duffing equation.  When we examined the latent dynamics for the experimental magnetic pendulum, we found very similar modes to the two previous systems (Fig. \ref{fig: eigenfunction 2}j-l). It is important to note that comparing system behavior, especially from observational data, is not straightforward using traditional local methods for dynamical systems analysis \cite{strogatz2018nonlinear}. However, with our approach, we automatically revealed the system’s behavior through the learned modes, allowing us to directly compare systems. Our results further suggest that we can make this comparison based on dimensionality, as systems with similar behavior are often modeled with the same number of dimensions (Fig. \ref{fig: studied systems}i-o). 

\captionsetup[figure]{labelformat=empty}
\subsection*{Empirical Lyapunov Stability}
\captionsetup[figure]{labelformat=empty}
\begin{figure}
    \centering
    \includegraphics[width=\textwidth]{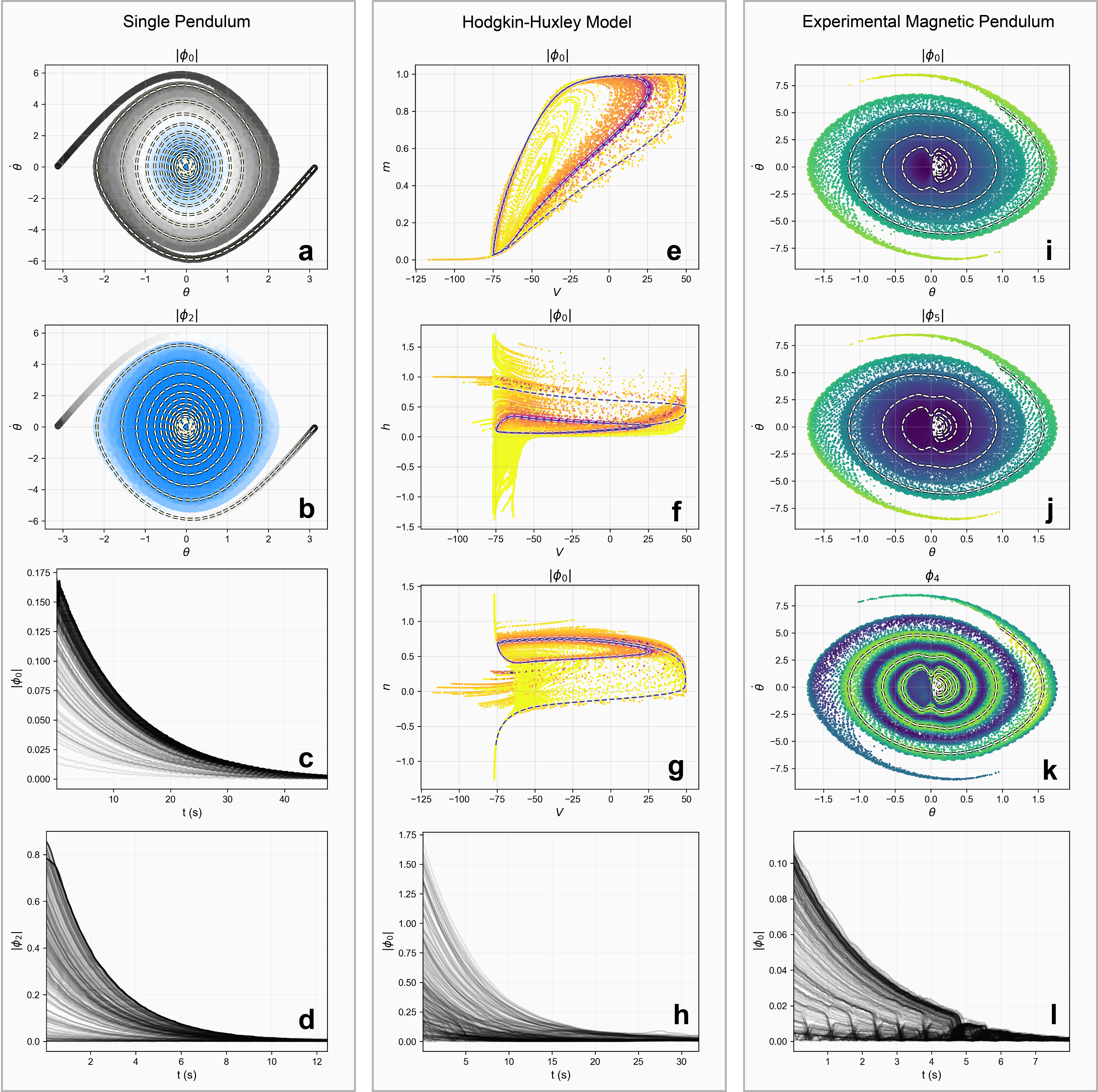}
        \caption{\textbf{Fig. 7 $\vert$ Lyapunov analysis and invariant sets.} \textbf{a} and \textbf{b}, A trajectory from the test dataset superimposed on two neural Lyapunov functions for the single pendulum. Each sublevel set for the neural Lyapunov functions serve as forward invariant sets. \textbf{c} and \textbf{d}, Empirical global stability analysis by evaluating ground truth trajectories using the neural Lyapunov function. This confirms the negative definite condition for the time rate of change of the Lyapunov function. \textbf{e-g}, A ground truth trajectory overlaid upon the neural Lyapunov function for the Hodgkin-Huxley limit-cycle attractor. \textbf{h}, Empirical verification of the Lyapunov function by evaluation with ground truth data. \textbf{i} and \textbf{j}, A trajectory superimposed upon two neural Lyapunov functions for the experimental magnetic pendulum. \textbf{k}, The same trajectory overlaid on the neural eigenfunction the corresponds to system's basin of attraction. \textbf{l}, Empirical stability analysis using trajectories from the experimental magnetic pendulum dataset.}
    \label{fig:lyapunov functions}
\end{figure}

 At large, stability analysis has been used for providing safety guarantees in chemical reactors, understanding the spread of disease, constructing controllers for autonomous systems, and designing aircraft wings. However, in most cases, stability analysis is limited to local regions in state space, making global stability analysis challenging for nonlinear systems. While Lyapunov's direct method provides a global approach, its practical application has been limited due to the difficulty of finding an appropriate Lyapunov function. In fact, there are no known general analytical methods to find Lyapunov functions. Our method, as noted earlier, automatically provides a neural Lyapunov function, enabling the application of Lyapunov stability analysis on observational data from nonlinear systems. We empirically analyzed the stability of the studied systems using this approach.  A Lyapunov function $V(\mathbf{x}) \in \mathbb{R}$ satisfies $V(\mathbf{x}) \geq 0$ and $\dot{V}(\mathbf{x}) \leq 0$ for all states $\mathbf{x} \in \mathbb{R}^n$ \cite{khalil2002control}. If the neural Lyapunov functions we've learned meet these conditions for observational data, we can conclude that the system is asymptotically stable. Moreover, each sub-level set of the function is forward invariant, which include static equilibria, more general attractors like limit cycles, and basins of attraction \cite{khalil2002control, deka2022koopman, mauroy2016global}. We constructed the neural Lyapunov functions by taking the magnitude of the eigenfunctions associated with eigenvalues having negative real parts.  The positive semi-definite condition is satisfied by taking the magnitude and the rate of change is negative due to the eigenvalue (\hyperref[sec:methods]{Methods}).  With this in mind, we used the learned eigenfunctions as powerful tools for stability analysis.

The superimposed time-series trajectories on the neural Lyapunov functions for the pendulum clearly showed evolution toward the minima, which coincided with the downward position of the pendulum at zero velocity (Fig. \ref{fig:lyapunov functions}a,b).  We evaluated the function on ground truth trajectories from the pendulum to verify the negative rate of change of the Lyapunov function and confirm that the downward position of the pendulum is globally asymptotically stable. These evaluated trajectories demonstrated exponential decay toward zero, certifying the automatically generated stability hypothesis (Fig. \ref{fig:lyapunov functions}c,d).

The neural Lyapunov function we learned for the Hodgkin-Huxley model revealed a particularly intricate structure when projected into the system’s state space (Fig. \ref{fig: eigenfunction 1}l-n).  The complexity of this representation is evident, especially considering that no analytical form of this function has yet been derived. By superimposing the function on trajectories from the system, we observed that our learned function clearly describes the system’s limit-cycle attractor (Fig. \ref{fig:lyapunov functions}e-g). Similar to our approach with the single pendulum, we verified the stability of this attractor by evaluating the Lyapunov function on trajectories from the system, confirming the exponential decay toward the attractor (Fig. \ref{fig:lyapunov functions}h).

We demonstrated our method’s ability to perform stability analysis on noisy experimental data using the magnetic pendulum. By plotting the system’s trajectories on the learned eigenfunctions, we emphasized their temporal evolution and identified equilibrium states at the minima (Fig. \ref{fig:lyapunov functions}i,j). The plotted trajectory remained entirely within the estimated basin of attraction, which we determined from the static mode of the dynamics (Fig. \ref{fig:lyapunov functions}k). We verified the asymptotic stability of the system's fixed points by analyzing trajectories from the system (Fig. \ref{fig:lyapunov functions}l). Similarly, the neural Lyapunov functions we generates for the other dynamical systems indicated the stability of their respective attractors (\hyperref[fig:extended data fig 3]{Extended Data Fig. 3}).

\section*{Discussion}
In this work, we introduced a novel framework for the automated global analysis and forecasting of nonlinear dynamical systems by learning low-dimensional linear embeddings directly from experimental data. Our approach stands out by discovering significantly lower-dimensional linear embeddings, often an order of magnitude smaller than those found by previous methods. Additionally, we achieved substantial improvements in long-horizon prediction accuracy and generalization performance. Unlike earlier methods that required refinement after learning, our approach directly produced neural Lyapunov functions, enabling straightforward global stability analysis. This was made possible through the integration of deep autoencoder networks, time-delay observables, and new regularization techniques. Using this framework, we derived nine novel coordinate representations for both prototypical and experimental dynamical systems, offering new insights into their behavior.

Future research could expand this method to process higher-dimensional data streams, such as video and audio. Enhancing the data efficiency of the framework is another path for improvement, particularly when data collection is costly or labor-intensive. Another promising direction is to extend this framework to controlled systems. Moreover, the current reliance on a brute force search to determine the appropriate embedding dimension suggests that future work could focus on developing methods for automatic embedding dimension discovery or uncovering theoretical connections between system behavior and latent linear dimensions.

\bibliography{references}

\section*{Methods}\label{sec:methods}
\subsection*{Koopman Operator Theory}Consider the following continuous-time dynamical system on a state space $\mathcal{M}\subseteq \mathbb{R}^n$
\begin{equation}
\begin{aligned}
    \dot{\mathbf{x}}= \mathbf{f}(\mathbf{x}),
\end{aligned}   
\end{equation}
where $\mathbf{x}(t)\in \mathcal{M}$ is the state at time $t\in\mathbb{R}$,  $\mathbf{f}: \mathcal{M} \rightarrow \mathcal{M}$ is the dynamics operator, and the overdot indicates the derivative with respect to time. The flow map $\mathbf{F}^t: \mathcal{M} \rightarrow \mathcal{M}$ integrates initial conditions $\mathbf{x}(t_0)\equiv \mathbf{x}_0$ to time $t$, and is defined as
\begin{equation}
    \begin{aligned}
        \mathbf{F}^t(\mathbf{x}_0) := \mathbf{x}_0 + \int_{t_0}^{t_0+t}\mathbf{f}(\mathbf{x}(\tau))d\tau\,.
    \end{aligned}
\end{equation}
Now, let us consider $\Psi(\mathcal{M})$ the set of scalar measurement or observable functions $\psi: \mathcal{M} \rightarrow \mathbb{C}$.  Then, the time-parameterized family of Koopman operators $\mathcal{K}^t: \Psi(\mathcal{M}) \rightarrow \Psi(\mathcal{M})$ is given by
\begin{equation}
    \begin{aligned}
        \mathcal{K}^t\psi(\mathbf{x}) = \psi \circ \mathbf{F}^t(\mathbf{x})\,.
    \end{aligned}
\end{equation}
If the flow, $\mathbf{F}^t$, is smooth and continuous we can define the infinitesimal generator $\mathcal{G}: \Psi(\mathcal{M}) \rightarrow \Psi(\mathcal{M})$ of $\mathcal{K}^t$ as
\begin{equation}
    \begin{aligned}
        \mathcal{G}\psi := \lim_{t\rightarrow 0^+} \frac{\mathcal{K}^t \psi - \psi}{t} = \lim_{t\rightarrow 0^+} \frac{\psi \circ \mathbf{F}^t - \psi}{t} = \dot{\psi}\,, 
    \end{aligned}
\end{equation}
and $\mathcal{K}^t = \textrm{exp}(\mathcal{G}t)$. The generator $\mathcal{G}$ gives rise to a continuous-time linear dynamical system in observable coordinates
\begin{equation}
    \begin{aligned}
        \dot{\psi} = \mathcal{G} \psi\,.
    \end{aligned}\label{eq: generator linear}
\end{equation}
When applying applying the chain-rule to $\dot{\psi}$ we also get that
\begin{equation}
    \begin{aligned}
        \dot{\psi}(\mathbf{x}) = \partial_{\mathbf{x}}\psi(\mathbf{x}) \dot{\mathbf{x}} = \partial_{\mathbf{x}}\psi(\mathbf{x}) \mathbf{f}(\mathbf{x})\,.
    \end{aligned}\label{eq: chain rule}
\end{equation}
An observable $\phi$ is an eigenfunction to the generator $\mathcal{G}$ if, for a corresponding eigenvalue $\mu \in \mathbb{C}$, 
\begin{equation}\label{eq: eigenfunc}
    \dot{\phi}(\mathbf{x}) = \mathcal{G}\phi(\mathbf{x}) = \mu\phi(\mathbf{x})\,.
\end{equation}
This also implies that $\phi$ is an eigenfunction for $\mathcal{K}^t$ with the eigenvalues  $\lambda^t = \textrm{exp}(\mu t)$. Eigenfunctions can also form a basis for observables with spectral decomposition
\begin{equation}
    \begin{aligned}
        \mathbf{\psi}(\mathbf{x}) = \sum_{j=1}^{\infty}\phi_j(\mathbf{x})\mathbf{v}_j\,,
    \end{aligned}\label{eq: infinite sum}
\end{equation}
where $\mathbf{v}_j$ is the $j$-th Koopman mode and $\mathbf{\psi}$ is a vector of observables. With the spectral decompostion (Eq. \ref{eq: infinite sum}), we can represent the flow of observables as
\begin{equation}
    \begin{aligned}
        \mathbf{\psi}(\mathbf{x}(t_k)) & = \mathcal{K}^{t_k}\sum_{j=1}^{\infty}\phi_j(\mathbf{x}_0)\mathbf{v}_j \\ 
        & = \sum_{j=1}^{\infty}\lambda_j^{t_k}\phi_j(\mathbf{x}_0)\mathbf{v}_j         \,.\\
    \end{aligned}\label{eq: kmd}
\end{equation}
which is known as Koopman Mode Decomposition \cite{mezic2005spectral}. Koopman Mode Decomposition highlights the benefits of the operator-theoretic perspective of dynamical systems, showing that a system's behavior is completely characterized by the interplay of $\lambda_j$, $\phi_j$, and $\mathbf{v}_j$. Lastly, the eigenfunctions of a dynamical system can be used to construct Lyapunov functions for stability analysis. A Lyapunov function $V(\mathbf{x}) \in \mathbb{R}$ satisfies $V(\mathbf{x}) \geq 0$ and $\dot{V}(\mathbf{x}) \leq 0$ for all $\mathbf{x}\in \mathcal{M}$. Then, the magnitude of eigenfunctions $|\phi|$ with $\textrm{Re}(\mu) \leq 0$ satisfy these conditions (Eq. \ref{eq: eigenfunc}). Moreover, any sub-level set 
\begin{equation}
    \mathcal{Q} =  \bigl\{ x \,|\, V(\mathbf{x}) \leq c \bigr\}
\end{equation}
with $c \geq 0 $ is forward invariant and the zero-level set is globally asymptotically stable \cite{deka2022koopman}. 
\subsection*{Loss Function}
Instead of considering the observables $\boldsymbol{\psi}$ as simple functions of the state, we input a small trajectory or set of delayed states. Time-delayed inputs have been widely used in a Koopman-informed models across various contexts and have proven to be a rich set of observables \cite{qian2022deep, takeishi2017learning, yuan2021flow, le2017higher, kamb2020time}. To illustrate this, we consider a vector of time-delayed states $\mathbf{X}_j$ starting at time $t_j$:
\begin{equation}
    \begin{aligned}
        \mathbf{X}_j =
        \begin{bmatrix}
            \mathbf{x}(t_j)^T, &  \mathbf{x}(t_j-\tau_0)^T, & \hdots, & \mathbf{x}(t_j-\tau_d)^T
        \end{bmatrix}^T\,,
    \end{aligned}
\end{equation}
where $\tau_i$ are delays and $i \in (0, 1, \dots, d)$.
We use the encoder $\boldsymbol{\psi}: \mathbb{R}^{(d+2)n} \rightarrow \mathbb{R}^{m}$ to transform the states into their latent representation in $\mathbb{R}^m$, and $\boldsymbol{\psi}^{-1}: \mathbb{R}^{m} \rightarrow \mathbb{R}^{n}$ to transform them back into state space, but only for the leading time $t_j$. In other words, the estimated or reconstructed state vector is given by $\hat{\mathbf{x}}_j = \boldsymbol{\psi}^{-1}(\boldsymbol{\psi}(\mathbf{X}_j))$. This reconstruction requirement gives us the first term of the loss function:
\begin{equation}\label{eq: recon loss}
    \begin{aligned}
        \mathcal{L}_{\mathbf{x}_0} = \frac{1}{n} \lVert \hat{\mathbf{x}}_0  - \mathbf{x}_0\rVert_2^2 = \frac{1}{n} \lVert \boldsymbol{\psi}^{-1}(\boldsymbol{\psi}(\mathbf{X}_0)) - \mathbf{x}_0 \rVert_2^2 \,.
    \end{aligned}
\end{equation}

In addition to reconstructing the leading input, the autoencoder and the latent space need to meet several more requirements informed by Koopman Operator Theory. First, by applying the chain rule to the observables (Eq. \ref{eq: chain rule}), we derive the second term in the loss function:
\begin{equation}\label{eq: chain-rule loss}
    \begin{aligned}
        \mathcal{L}_{\dot{\boldsymbol{\psi}}} = \frac{1}{m} \lVert \mathcal{G} \boldsymbol{\psi}(\mathbf{X}_0) - \partial_{\mathbf{X}} \boldsymbol{\psi}(\mathbf{X}_0) \dot{\mathbf{X}}_0  \rVert_2^2 \,.
    \end{aligned}
\end{equation}
Next, to ensure good predictions with the linear model in latent space, we include the following term in the loss function:
\begin{equation}\label{eq: latent pred loss}
    \begin{aligned}
        \mathcal{L}_{\mathbf{\psi}_t} = \frac{1}{mT} \sum_{j=1}^{T} \gamma^{j-1}\lVert \hat{\boldsymbol{\psi}}(\mathbf{X}_j) - \boldsymbol{\psi}(\mathbf{X}_j)  \rVert_2^2 =  \frac{1}{mT} \sum_{j=1}^{T} \gamma^{j-1} \lVert e^{\mathcal{G}(t_j-t_0)} \boldsymbol{\psi}(\mathbf{X}_0) - \boldsymbol{\psi}(\mathbf{X}_j)  \rVert_2^2 \,.
    \end{aligned}
\end{equation}
Here, the matrix exponential is the analytical solution to the linear dynamics, integrating initial conditions from time $t_0$ to $t_j$. The sum in the latent space prediction loss accounts for future predictions over a time horizon $T$ (Eq. \ref{eq: latent pred loss}). It's worth note that computing the prediction loss is computationally inexpensive. In practice, we can form a stack of time-scaled matrix exponentials (i.e., $e^{\mathcal{G}(t_j-t_0)}$) and multiply them by a batch of latent vectors $\boldsymbol{\psi}(\mathbf{X}_0)$ to predict a batch of future latent embeddings. This method is far more efficient than performing multi-step predictions for nonlinear systems due to its use of recurrence relationships.

\begin{figure}
    \centering
    \includegraphics[width=\textwidth]{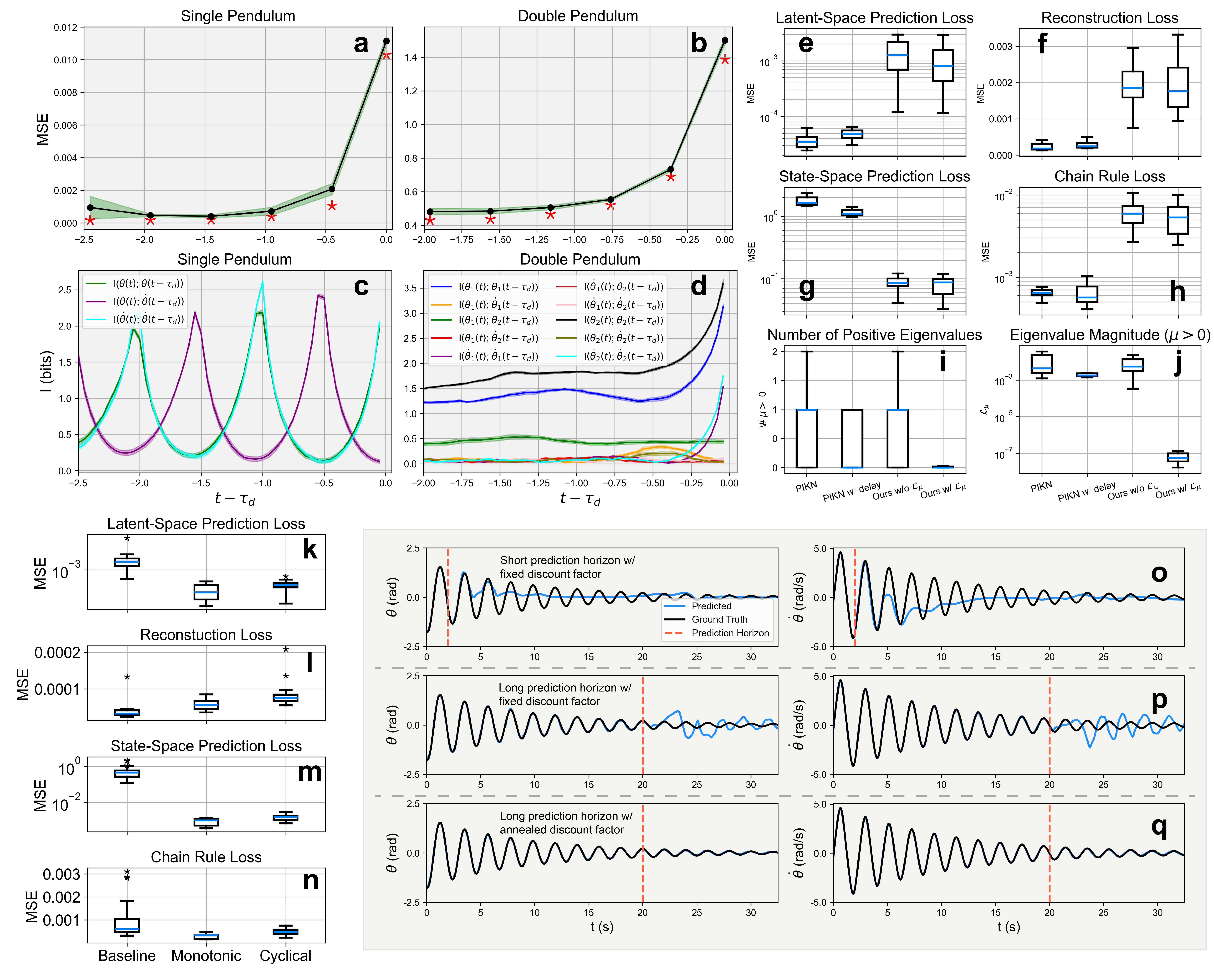}
    \caption{\textbf{Extended Data Fig. 1 $\vert$ Baseline Comparisons.} \textbf{a}, The effect of using time-delay on the long-horizon prediction error of the single pendulum.  \textbf{b}, The effect of using time-delay on the long-horizon prediction error of the chaotic double pendulum. \textbf{c}, The mutual information between the time-delayed states in the single pendulum. One cycle of mutual information indicates sufficient time-delay for accurate long-horizon predictions. \textbf{d}, The mutual information between time-delayed states in the double pendulum, revealing a sufficient delay for accurate predictions after the curves reach a local minima. \textbf{e-j}, Comparison of our method with the closest related work, Physics Informed Koopman Network (PIKN) \cite{liu2022physics}. The PIKN lacks the state space prediction loss term (Eq. \ref{eq: state pred loss}), and the eigenvalue penalty loss (Eq. \ref{eq: G reg}) resulting in a collapse of the latent space, poor state space predictions, and false eigenvalues. \textbf{k-n}, Comparison of using our hyperparameter annealing strategies against baseline random search. Monotonic and Cyclical annealing greatly improve long-horizon predictions. \textbf{o-p}, Predicted and ground truth trajectories for models trained with different prediction horizons. \textbf{o}, Models trained with a modest horizon but fixed discount factor resulted in fair generalization for a short time after training horizon. \textbf{p} Models trained with an extended-horizon and fixed discount factor overfit to the training horizon. \textbf{q}, Models trained with a long-horizon and varying discount factor achieved prediction generalization.}
    \label{fig:extended data fig 1}
\end{figure}

\begin{figure}
    \centering
    \includegraphics[width=\textwidth]{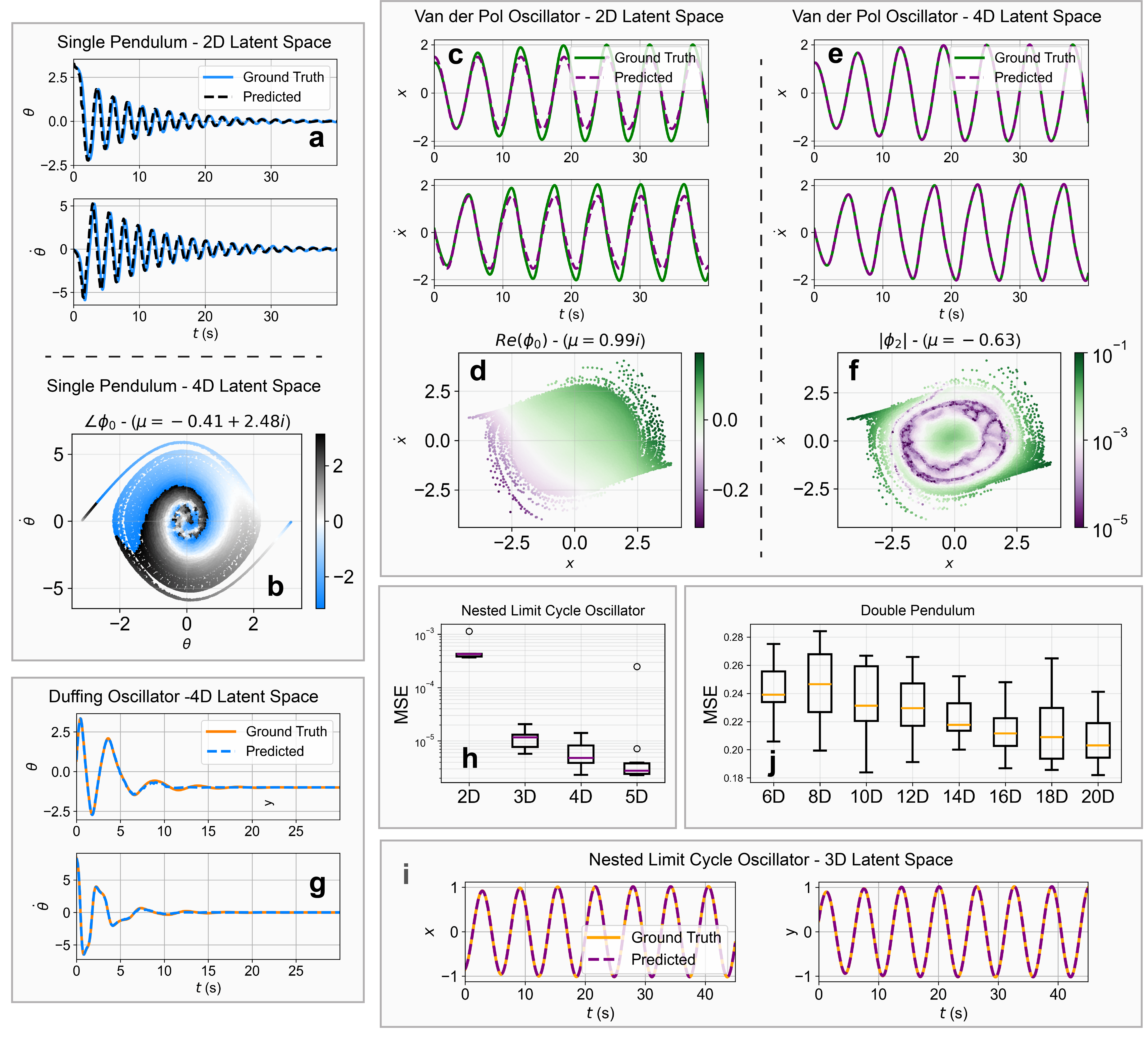}
    \caption{\textbf{Extended Data Fig. 2 $\vert$ Predictions in Various Dimensions.} \textbf{a}, Predictions for the single pendulum in 2D cannot capture frequency shifting. \textbf{b}, The model for the single pendulum in 4D predicts additional frequency content and an eigenfunction that is unnecessary to perform long-horizon predictions. \textbf{c} and \textbf{d}, The 2D model for the Van der Pol oscillator failed to predict transient behavior in the system and had an asymmetric latent space. \textbf{e} and \textbf{f}, The 4D model for the Van der Pol oscillator achieved accurate predictions but contained nonessential unstructured modes. \textbf{g}, Predictions from the 4D model for the Duffing oscillator are accurate except for low-amplitude intra-well oscillations. \textbf{h}, Box and whisker plots for the MSE of the nested limit-cycle oscillator across latent dimensions showing a large dropoff from 2D to 3D. \textbf{i}, Predicted and ground truth trajectories for a model for the nested limit-cycle system trained with a 3D latent space. \textbf{j}, Box and whisker plots for the MSE of double pendulum models across dimensions showing little correlation between error and dimension due to the chaotic nature of the system.}
    \label{fig:extended data fig 2}
\end{figure}

\begin{figure}
    \centering
    \includegraphics[width=\textwidth]{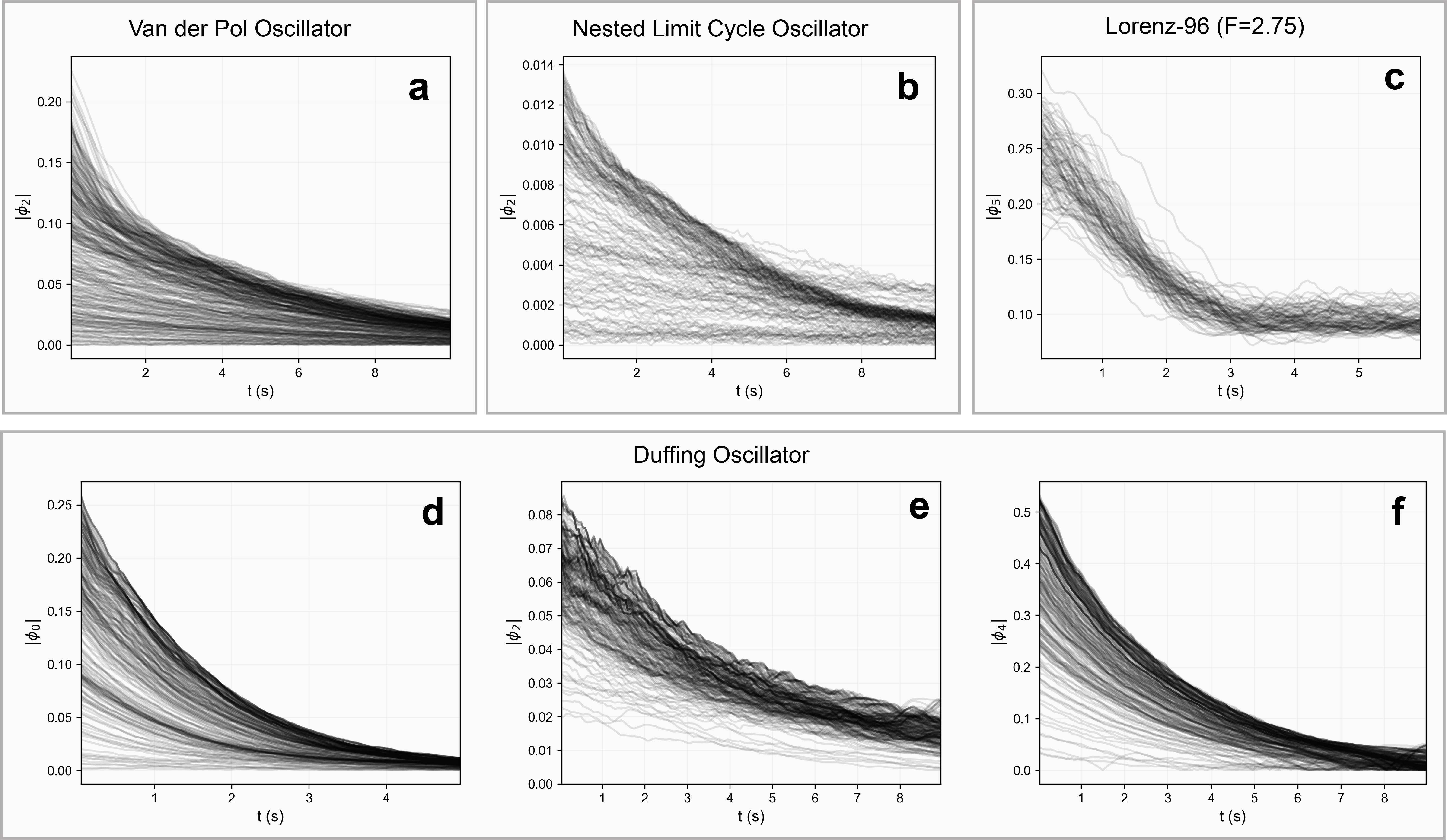}
    \caption{\textbf{Extended Data Fig. 3 $\vert$ Empirical Stability analysis.} \textbf{a-f}, Learned Lyapunov functions for the studied systems evaluated on test trajectories.}
    \label{fig:extended data fig 3}
\end{figure}

In theory, if the terms presented so far are sufficiently minimized (Eq. \ref{eq: recon loss}-\ref{eq: latent pred loss}), future predictions in state space should also perform well. In fact, without the time-delay, these terms form the loss function for Physics Informed Koopman Networks (PIKNs) (Eq. \ref{eq: recon loss}-\ref{eq: latent pred loss}) \cite{liu2022physics}. However, not explicitly enforcing prediction quality in state space can result in a shrinking latent space during optimization, where the prediction loss (Eq. \ref{eq: latent pred loss}) could be minimized by allowing the encoder to output zero for every input. To avoid such shortcuts, we introduced an additional term that directly quantifies the quality of predictions in state space:
\begin{equation}\label{eq: state pred loss}
    \begin{aligned}
        \mathcal{L}_{\mathbf{x}_t} = \frac{1}{nT} \sum_{j=1}^{T}\gamma^{j-1} \lVert \hat{\mathbf{x}}_j - \mathbf{x}_j  \rVert_2^2 =   \frac{1}{nT} \sum_{j=1}^{T} \gamma^{j-1} \lVert \boldsymbol{\psi}^{-1}(e^{\mathcal{G}(t_j-t_0)} \boldsymbol{\psi}(\mathbf{X}_0)) - \mathbf{x}_j  \rVert_2^2 \, .
    \end{aligned}
\end{equation}
We replaced the 2-norm in the state space prediction loss with the Mahalanobis distance with diagonal covariance for the Hodgkin-Huxley dataset to account for the scaling of the states. Lastly, we directly penalized the spectrum of $\mathcal{G}$ to encourage the learning of stable or neutrally stable dynamics with the following term:
\begin{equation}\label{eq: G reg}
    \begin{aligned}
        \mathcal{L}_{\mu} = \sum_{j=0}^{m}\textrm{max}(0,\textrm{Re}(\mu_{j}))\,.
    \end{aligned}
\end{equation}
The total loss is then given by:
\begin{equation}\label{eq: total loss}
    \begin{aligned}
        \mathcal{L} = \mathcal{L}_{\mathbf{x}_0}+\alpha_1\mathcal{L}_{\dot{\boldsymbol{\psi}}}+\alpha_2\mathcal{L}_{\boldsymbol{\psi}_t}+\alpha_3\mathcal{L}_{\mathbf{x}_t}+\alpha_4\mathcal{L}_{\mu}\,.
    \end{aligned}
\end{equation}
To assess the impact of the additional loss terms (Eq. \ref{eq: state pred loss},\ref{eq: G reg}) on the performance of the learned model and compare our approach to PIKNs, we conducted ablation experiments using simulated data from the magnetic mass-spring-damper system. We selected this system because of its demonstration of nonlinear behavior and global stability, implying that the learned eigenvalues should be negative. In these experiments, we assessed several metrics, including latent space prediction error, state space prediction error, reconstruction error, chain rule loss error, and the number and magnitude of positive eigenvalues (\hyperref[fig:extended data fig 1]{Extended Data Fig. 1}e-j). 

In the first trials, we examined PIKNs, which struggled with predicting future states, assessing stability, and maintaining a stable latent space. In subsequent trials, we augmented the PIKN model with time-delayed inputs, which marginally improved both stability and state space prediction capabilities. We then introduced the state space prediction loss term alongside the time-lagged inputs, which significantly enhanced the accuracy of state space predictions by an order of magnitude compared to the original PIKN models. However, despite these improvements, the models still failed to accurately estimate stability, with a median of one positive eigenvalue per model. Ultimately, the introduction of the eigenvalue penalty (Eq. \ref{eq: G reg}) resulted in the best performance.
\captionsetup[figure]{labelformat=empty}

\captionsetup[figure]{labelformat=empty}

\subsection*{Time-Delay Selection}
Using time-delayed data to analyze or model a dynamical systems is an established approach. In fact, Taken's theorem, first published in 1981, demonstrated that attractors can be reconstructed from time-delayed partial state measurements \cite{takens1981detecting}. Since, numerous methods have been proposed for modelling and forecasting dynamics from time-delayed measurements including local polynomial methods \cite{abarbanel1994predicting}, time-delayed or Hankel DMD \cite{yuan2021flow, le2017higher, brunton2017chaos, kamb2020time}, and deep learning-based models \cite{takeishi2017learning, bakarji2022discovering, qian2022deep}.

Regardless of the method used, the length of time-delay is an important hyperparameter that significantly impacts model performance (\hyperref[fig:extended data fig 1]{Extended Data Fig. 1}a,b). Information theory has been suggested in previous work as a useful tool for selecting the length of time-delay \cite{fraser1986independent, fraser1989information, abarbanel1994predicting, abarbanel2012analysis}. However, much of this research is focused on attractor reconstruction as opposed to Koopman-informed modelling. Furthermore, few studies, if any, have explored the relationship between model performance and information-theoretic metrics. In this section, we examine the relationship between mutual information (MI) and future state prediction error. From this relationship, we propose a principled approach to selecting a lower bound on the length of time-delay for latent linear models.

The MI in bits between two discrete random variables $A$ and $B$ is given by:
\begin{equation}
    \begin{aligned}
        I(A; B) = \sum _{a\in \mathcal{A} } \sum_{b \in \mathcal{B}} P_{(A, B)} (a, b)\, \textrm{log}_2 \left( \frac{P_{(A, B)} (a, b)}{P_{A} (a) P_{B} (b)} \right)
    \end{aligned}
\end{equation}
where $P_{(A, B)}$ is the joint probability mass function, and $P_{A}$, $P_{B}$ are the marginal probability mass functions for $A$ and $B$, respectively. In the context of collected trajectories from a dynamical system, we aim to calculate the MI between every, or many, combinations of states and their time-delayed counterparts. For example, the MI between a set of states at an arbitrary time $x_a(t_j)$ and another state at a previous instant $x_b(t_j-\tau_d)$ is expressed as:
\begin{equation}
        I(x_a(t); x_b(t-\tau_d)) = \sum _{t_j} P(x_a(t_j), x_b(t_j-\tau_d))\, \textrm{log}_2 \left( \frac{P(x_a(t_j), x_b(t_j-\tau_d))}{P(x_a(t_j)) P(x_b(t_j-\tau_d))} \right)\,.
\end{equation}
Naturally, if $a=b$, this measures the mutual information between a state at an arbitrary instant and a previous instant. 

We calculated the mutual information for a single pendulum (\hyperref[fig:extended data fig 1]{Extended Data Fig. 1}c). The plot shows the mutual information vs. time delay ($t-\tau_d$) for each combination of the states $\theta$, and $\dot{\theta}$. We observe a nearly periodic pattern in $I$ as a function of the delay. Moreover, the mutual information between states and themselves is out of phase with mutual information between $\theta$ and $\dot{\theta}$. Similarly, we calculated the mutual information between each combination of states in a double pendulum (\hyperref[fig:extended data fig 1]{Extended Data Fig. 1}d). Due to the chaotic dynamics, the mutual information did not exhibit a periodic pattern as a function of time delay, nor did the decrease in information occur monotonically. Interestingly, we observed an increase in mutual information between the angular positions and their velocities at approximately a half-second delay.

When we consider the prediction error as a function of delay alongside the time-delayed mutual information, we observe that predictions reach a point of diminishing returns when plotted against $t-\tau_d$. In the case of the single pendulum, this point coincides with one full-period in the mutual information signal, suggesting that for periodic systems, one full period of information should be included in the model. We also observed a relationship between the mutual information in the double pendulum and the predictive ability of the trained models. The prediction error appeared to plateau once each of the mutual information curves reached a local minimum. For instance, while the blue and grey mutual information curves reached a local minima at approximately $t-\tau_d=0.6s$, the yellow and olive curves did not reach a minima until $t-\tau_d=1.25s$, which coincided with the point of diminishing returns (\hyperref[fig:extended data fig 1]{Extended Data Fig. 1}d).

With these results in mind, we can use the mutual information between the states in a system to approximate the model's predictive capability as a function of time delay. This provides us with an estimate for the lower bound of the time delay, and this process is significantly faster than performing hyperparameter search or running controlled experiments.

\subsection*{Annealing Strategy for the Loss Function and Discount Factor}
Minimizing the total loss (Eq. \ref{eq: total loss}) presents several challenges. First, the loss landscape for a function with this many terms is likely filled with local extrema, making global optimization difficult. Second, the outcome of the optimization heavily depends on the values of the weights of each loss term, $\alpha$, and the discount factor $\gamma$. Many researchers address a weighted multi-parametric loss function with random search or hand-tuning. While this type of brute force search can still yield an adequate model, it does not improve training stability or model generalization and is often computationally expensive. 

The practice of annealing, which entails systematically adjusting the hyperparameters during training, is widely recognized for improving training stability and enhancing model generalization. This technique is most frequently applied to the learning rate. For instance, Smith et.al. proposed oscillating learning rates in a periodic or quasi-periodic pattern to balance exploration and exploitation during training \cite{smith2017cyclical, smith2019super}. Recent work on variational autoencoders (VAEs) has shown that cyclically annealing the coefficients in the loss function can improve learning by mitigating vanishing of the Kullback-Liebler divergence \cite{fu2019cyclical}. 

In accordance with Fu et.al., we set the $\alpha$ coefficients in the loss function (Eq. \ref{eq: total loss}) with the following schedule:
\begin{equation}\label{eq: annealing}
    \begin{aligned}
        \alpha^i = 
        \begin{cases}
        g(k), & k \leq R \\
        1, & k > R
        \end{cases}
        \quad \textrm{with} \quad k = \frac{\textrm{mod}(i-1, T_c/M)}{T_c/M}\, ,
    \end{aligned}
    \
\end{equation}
where $\alpha^i$ represents the hyperparameter value at the $i^{th}$ training iteration  \cite{fu2019cyclical}. We define integers $M$ as the number of cycles and $T_c$ as the total number of training iterations. The parameter $R \in [0, 1]$ indicates the proportion of each cycle during which $\alpha$ remains at its maximum value of 1, and $g$ is a function that increases monotonically from 0 to 1. For simplicity, we consider linear or sigmoid functions for $g$. Importantly, training invariably concludes with $\alpha$ set to 1.

\begin{figure}
    \centering
    \includegraphics[width=\textwidth]{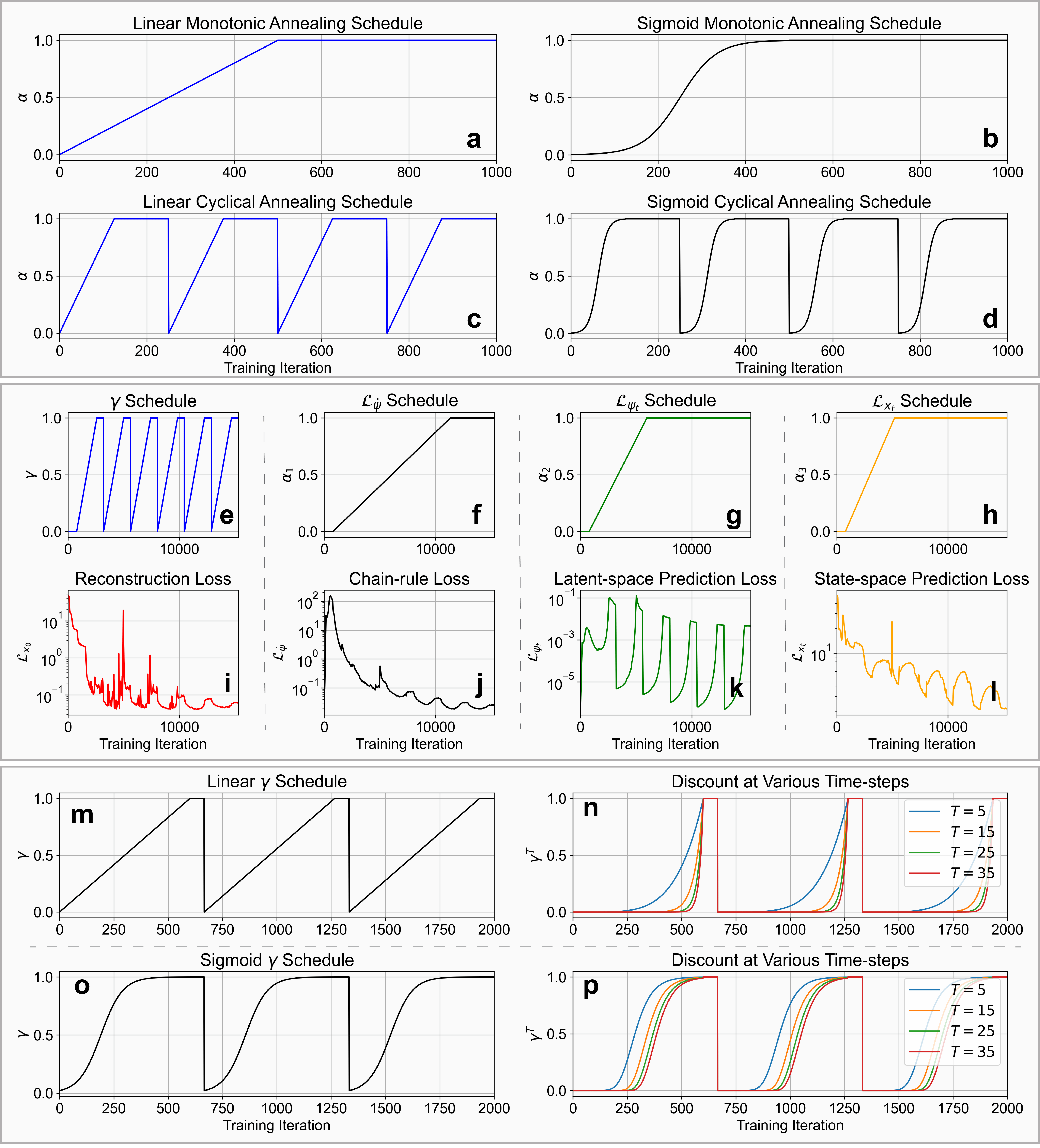}
    \caption{\textbf{Extended Data Fig. 4 $\vert$ Hyperparameter annealing.} \textbf{a}, Example linear monotonic annealing schedule used for the coefficients of the loss function and the discount factor. \textbf{b}, Sigmoid monotonic annealing schedule. \textbf{b}, Sigmoid monotonic annealing schedule. \textbf{c}, Linear cyclical annealing schedule. \textbf{d}, Sigmoid cyclical annealing schedule. \textbf{e-l}, Sample annealing and training loss curves that illustrate the challenges with linear cyclical annealing. The rapid increase in the discount factor (\textbf{e}) induces exploding gradients at around 5000 training iterations (\textbf{i}). \textbf{m} and \textbf{n}, The nonlinear application of the discount factor results results in a binary weighting of predictions at longer training horizons $T$. \textbf{o} and \textbf{p} Sigmoid annealing results in a smoother annealing of the prediction horizon.}
    \label{fig:extended data fig 4}
\end{figure}

In this study, we categorize the annealing process into two types: when $M=1$, we refer to it as monotonic or standard annealing, and when $M>1$, we call it cyclical annealing. The schedule parameters $M$ and $R$ can vary across different coefficients in the loss function. Notably, we do not apply annealing to the reconstruction loss, as we regard reconstruction as fundamental to the model's effectiveness throughout training. Inadequate reconstruction quality would adversely impact the model’s capability to make accurate future predictions in state space. To ensure accurate reconstruction, we warm-start or pre-train each network on the reconstruction loss alone. Examples of both cyclical and standard annealing schedules, using linear and sigmoid functions, demonstrate how they vary over the course of training iterations (\hyperref[fig:extended data fig 4]{Extended Data Fig. 4}).
\captionsetup[figure]{labelformat=empty}

In addition to the $\alpha$ parameters in the loss function, the discount factor and prediction horizon have a disproportionate impact on model training, and generalization performance. Many learning methods opt to optimize for single or few time-step predictions. This approach can be effective if the loss is sufficiently small, as the simple objective has a regularization effect. However, in many cases, these models are prone to underfitting, as we will show. On the other hand, optimizing for multiple time-step predictions comes with it's own challenges such as overfitting and exploding gradients. To address these issues, we use discount factor annealing as a method to balance time-horizon trade-offs and improve model performance.

While annealing the different terms in the loss function can be viewed as altering the importance of each term throughout training, discount factor annealing effectively changes the prediction horizon. Gradually increasing the discount factor over time acts as temporal curriculum learning. In other words, the model must first learn to make short-term predictions before progressions to longer-term predictions. 

To further motivate the need for discount factor annealing, we examined the model predictions for a pendulum, as depicted in (\hyperref[fig:extended data fig 1]{Extended Data Fig. 1}o-q). This figure contrasts three models that differ solely in their prediction horizon or discount factor. First, we compared the predicted and actual time series when using a model with a relatively short prediction horizon, where each prediction step is weighted equally with $\gamma=1$ (\hyperref[fig:extended data fig 1]{Extended Data Fig. 1}o). We observed that the predictions remain accurate up to, and slightly beyond, the training horizon, after which they deteriorate.

A straightforward strategy to improve the model's predictive accuracy might be to extend the prediction horizon. While extending the horizon does improve the model's predictive performance within the training horizon, its ability to generalize beyond this point is limited and nonphysical (\hyperref[fig:extended data fig 1]{Extended Data Fig. 1}p). Thus, training with longer prediction horizons may lead to overfitting. Finally, after implementing a cyclical annealing schedule for $\gamma$, we observe that the model is able to generalize well past the training horizon (\hyperref[fig:extended data fig 1]{Extended Data Fig. 1}q). These results also illustrate the importance of using a longer validation horizon than training horizon during model selection. Even if a model performs well over many time-steps, it may still fail to generalize to extended horizons.

A high-level overview of the loss curves when employing linear monotonic annealing for the loss coefficients $\alpha$, and linear cyclical annealing for the discount factor provides insight into the training dynamics (\hyperref[fig:extended data fig 4]{Extended Data Fig. 4}e-l). These loss curves highlight the challenges associated with annealing the discount factor while training with extended prediction horizons. Notably, we observed a significant uptick in the loss metrics at the 5000th iteration (\hyperref[fig:extended data fig 4]{Extended Data Fig. 4}i-l). This increase in the loss aligns with the discount factor elevating to a value of 1, suggesting a correlation between the rate of change of the discount factor and a temporary regression of the model performance.

When we examined the relationship between the discount factor $\gamma$ and its application over various future time steps, $\gamma^T$, we uncovered distinct behaviors depending on the chosen annealing function (\hyperref[fig:extended data fig 4]{Extended Data Fig. 4}m-p). A linear annealing function showed a pronounced escalation in the value of $\gamma^T$ as $T$ increases, particularly noticeable at extended horizons. This change become starkly apparent even at relatively short horizons, such as $T=25$, where the weighting applied to those predictions essentially becomes binary (\hyperref[fig:extended data fig 4]{Extended Data Fig. 4}n). This phenomenon is attributed to the fact that, under a linear framework, the effective horizon remains small until $\gamma$ approaches $1$, a consequence of the exponential decay inherent in the geometric series used for discounting. So, the impact of the annealing function on the effective prediction horizon and the overall training dynamics should not be underestimated. 

To resolve the abrupt change in weighting at long prediction horizons observed with linear annealing, we adopted a sigmoid function for the main trials in this work. Using a sigmoid function ensured a more gradual adjustment in the weights assigned to time steps farther into the future compared to linear annealing (\hyperref[fig:extended data fig 4]{Extended Data Fig. 4}p). By incrementing the discount factor according to a sigmoid curve, we extended the effective prediction horizon in a more gradual and controlled manner and mitigated the binary weighting effect seen with linear annealing.
\captionsetup[figure]{labelformat=empty}
Lastly, we performed experiments to evaluate the efficacy of different annealing strategies. We compared monotonic and cyclical sigmoid annealing strategies against a baseline method, which employed simple random search to determine the coefficients of the loss function and the discount factor. For each strategy— monotonic, cyclical, and baseline— we set the training and testing horizons to $T=400$ and $T=600$, respectively. In the baseline approach, we sampled the discount factor from a uniform distribution with a range of $0.9$ to $1.0$, allowing for variations in the effective prediction horizon. As previously noted, we fixed the the coefficient for reconstruction loss at $1$, whereas we uniformly sampled the coefficients $\alpha_i$  from a range of $0.01$ to $1.0$. The detailed hyperparameter settings for the annealing experiments are documented in Table \ref{tab: Studies table 4: annealing ablation MI} and \ref{tab: annealing table}.

We observed significant enhancements in model performance with both cyclical and monotonic annealing when compared to the baseline method, particularly in forecasting accuracy (\hyperref[fig:extended data fig 4]{Extended Data Fig. 4}k-n). Moreover, we observed nearly two orders of magnitude improvement in both latent-space and state-space prediction error, with minimal to no compromise on $\mathcal{L}_{x_0}$ and $\mathcal{L}_{\dot{\psi}}$. We note a slight edge in performance for monotonic over cyclical. For this reason, we choose to employ monotonic for the models trained in this work. For especially long training horizons, we increase the time step for predictions to achieve extended forecasts without destabilizing training from a large $T$.

\subsection*{Data Collection and Model Training}
We generated the simulated datasets in Python with the SciPy library using 4th order Runge-Kutta numerical integration. The number of trajectories we collected for each dataset, along with the trajectory length, size of the time-step, etc. are given in Table \ref{tab: Studies table 1}-\ref{tab: Studies table 4: annealing ablation MI}. We sampled initial conditions for the dissipative systems from a beta distribution to prevent bias in the datasets towards low-energy trajectories. The dataset for the magnetic-mass-spring-damper was balanced to contain an equal number of trajectories in each attractor. We trained the neural networks using Pytorch \cite{paszke2019pytorch} and Lightning \cite{falcon2019pytorch}. We used the AdamW optimizer and OneCycleLR for all of the models  and performed hyperparameter search with Optuna \cite{loshchilov2017decoupled, smith2019super, akiba2019optuna}. The various training and computational hardware details we used for each experiment is given in Table \ref{tab: Studies table 1}-\ref{tab: Studies table 4: annealing ablation MI}. We used a combination of the Moteus r4.11 controller and the Moteus mj5208 brushless motor as a servomotor to collect data for the experimental pendulum setups. We modeled the simulated systems according the equations given in the Supplementary Materials.





\section*{Acknowledgments}
This work was supported by the National Science Foundation Graduate Research Fellowship, the ARL STRONG program under awards W911NF2320182
and W911NF2220113, by ARO W911NF2410405, by DARPA FoundSci program under
award HR00112490372, and DARPA TIAMAT program under award HR00112490419.

\newpage
\section*{Supplementary Materials}
\setcounter{figure}{0} 
\setcounter{table}{0}  
\setcounter{equation}{0} 
\renewcommand{\thefigure}{S\arabic{figure}} 
\renewcommand{\thetable}{S\arabic{table}}  
\renewcommand{\theequation}{S\arabic{equation}} 
\subsection*{A. Simulated Dynamical Systems}\label{sec:supplementary materials a}
We modeled the simulated systems according the following equations. We constructed the Van der Pol oscillator dataset with the following ordinary differential equation:
\begin{equation}
    \begin{aligned}
        \ddot{x} = \nu (1 - x^2) \dot{x} - x\,,
    \end{aligned}\label{eq: vdp eom}
\end{equation}
with $\nu > 0$. The second-order differential equation we used to describe dynamics of the Duffing oscillator was given by:
\begin{equation}
    \begin{aligned}
        \ddot{x} =  x + \beta \dot{x} - x^3\,,
    \end{aligned}\label{eq: duffing eom}
\end{equation}
where $\beta=0.5$. We generated the Hodgkin-Huxley data with the following four ordinary differential equations:
\begin{equation}
    \begin{aligned}
        \dot{V} &= \frac{I_{\text{ext}} - g_{\text{Na}} m^3 h (V - E_{\text{Na}}) - g_{\text{K}} n^4 (V - E_{\text{K}}) - g_{\text{L}} (V - E_{\text{L}})}{C_m} \\
        \dot{m} &= \alpha_m(V) (1 - m) - \beta_m(V) m \\
        \dot{h} &= \alpha_h(V) (1 - h) - \beta_h(V) h \\
        \dot{n} &= \alpha_n(V) (1 - n) - \beta_n(V) n\,.
    \end{aligned}\label{eq: HH0}
\end{equation}
Where $V$ is the membrane potential, $m$ is the probability of sodium channel activation, $h$ is the probability of sodium channel inactivation, and $n$ is probability of potassium channel activation. The constant parameters we used in Eq. \ref{eq: HH0} are listed in Table \ref{tab: HH params}, while the voltage dependant parameters used are given by:
\begin{equation}
    \begin{aligned}
        \alpha_m(V) &= \frac{0.1 (V + 45)}{1 - \exp\left(-\frac{V + 45}{10}\right)} \\
        \beta_m(V) &= 4.5 \exp\left(-\frac{V + 70}{18}\right) \\
        \alpha_h(V) &= 0.07 \exp\left(-\frac{V + 70}{20}\right) \\
        \beta_h(V) &= \frac{1}{1 + \exp\left(-\frac{V + 40}{10}\right)} \\
        \alpha_n(V) &= \frac{0.01 (V + 60)}{1 - \exp\left(-\frac{V + 60}{10}\right)} \\
        \beta_n(V) &= 0.15 \exp\left(-\frac{V + 70}{80}\right)\,.
    \end{aligned}\label{eq: HH1}
\end{equation}

We created the Lorenz 96 dataset with the following dynamical system. Its dynamics are described by $N$ states representing the value of an atmospheric quantity across $N$ sections of a latitude with $N \geq 4 $. The dynamics of the quantity in the $i$th section, $x_i$, is governed by:
\begin{equation}
\begin{aligned}
    \frac{d x_i}{dt} = (x_{i+1}-x_{i-2})x_{i-1} - x_i + F\,,
\end{aligned}
\end{equation}
where $F$ is external forcing. The boundary conditions that provide continuity across the ends of the latitude circle are:
\begin{equation}
\begin{aligned}
    x_{-1} = x_{N-1}, \quad x_{0} = x_{N}, \quad x_{1} = x_{N+1}\,.
\end{aligned}
\end{equation}
\begin{table}[h]
\centering
\begin{tabular}{c|c|c}
\textbf{Parameter} & \textbf{Value} & \textbf{Description} \\
\hline
$C_m$ & 1.0 & Membrane capacitance, in $\mu$F/cm$^2$ \\
$g_{\text{Na}}$ & 120.0 & Maximum sodium conductance, in mS/cm$^2$ \\

$g_{\text{K}}$ & 36.0 & Maximum potassium conductance, in mS/cm$^2$ \\

$g_{\text{L}}$ & 0.3 & Maximum leak conductance, in mS/cm$^2$ \\

$E_{\text{Na}}$ & 50.0 & Sodium equilibrium potential, in mV \\

$E_{\text{K}}$ & -77.0 & Potassium equilibrium potential, in mV \\

$E_{\text{L}}$ & -54.387 & Leak equilibrium potential, in mV \\

$I_{\text{ext}}$ & 10.0 & External current, in $\mu$A/cm$^2$ \\
\end{tabular}
\caption{Parameters used in the Hodgkin-Huxley model}
\label{tab: HH params}
\end{table}
Our simulated pendulum dataset was generated with the following point-mass model
\begin{equation}\label{eq: pend eom}
    \begin{aligned}
        \ddot{\theta} = -\frac{g}{l}\sin(\theta) - \beta\dot{\theta}\,.
    \end{aligned}
\end{equation}
Here, $\theta$ is the angular position of the pendulum relative to vertical, and the parameters details are given in Table \ref{tab: single pend tab}. The model we used to simulate the magnetic-mass-spring-damper was introduced in \cite{wang2022model} and is given by:
\begin{equation}
    \begin{aligned}
        m\ddot{x} + c\dot{x} + kx = \alpha(x-b)\left[ 12h^2 - 3(x-b)^2\right]\left[h^2 + (x-b)^2\right]^{-7/2}\,.
    \end{aligned}\label{eq: mmsd eom}
\end{equation}
The horizontal translation of the magnet is given by the state $x$ and the details of the parameters are given in Table \ref{tab: mmsd tab}. The model we used to simulate the nested limit cycle oscillator is given the following two-dimensional dynamical system:
\begin{equation}
    \begin{aligned}
        \dot{x} &= x \left( \frac{1}{3} - r \right) \left( \frac{2}{3} - r \right) (1 - r) + y \\
        \dot{y} &= y \left( \frac{1}{3} - r \right) \left( \frac{2}{3} - r \right) (1 - r) - x \\
        r &= \sqrt{x^2 + y^2}\,.
    \end{aligned}
\end{equation}
Lastly, the we used this model of a point mass double pendulum to create a synthetic dataset for the mutual information experiments:
\begin{equation}\label{eq: double pend eom}
    \begin{aligned}
        \ddot{\theta}_1 & = \frac{1}{l_1(m_1 + m_2\sin^2(\theta_1 - \theta_2))}\Big[-m_2\sin^2(\theta_1 - \theta_2)\big( l_1\dot{\theta}_1^2 \cos(\theta_1 -\theta_2) + l_2 \dot{\theta}_2^2 \big) \\ & + m_2 g \sin(\theta_2)\cos(\theta_1 - \theta_2) - (m_1+m_2) g \sin(\theta_1) - \beta\dot{\theta}_1 \Big]\,, \\ 
        \ddot{\theta}_2 & = \frac{1}{l_2(m_1 + m_2\sin^2(\theta_1 - \theta_2))}\Big[ (m_1+m_2)\big(l_1\dot{\theta_1}^2 \sin(\theta_1 - \theta_2) - g\sin(\theta_2) \\ 
        & + g \sin(\theta_1)\cos(\theta_1 - \theta_2) \big)  + m_2 l_2 \dot{\theta_2}^2\sin(\theta_1 - \theta_2)\cos(\theta_1 - \theta_2) - \beta\dot{\theta}_2 \Big]\,.
    \end{aligned}
\end{equation}
Here, $\theta_1$ is the angle of the top link and the $\theta_2$ the angle of the bottom. A description and the values of parameters we used in the model is given in Table \ref{tab: double pend table}.
\begin{table}[h]
\centering
\begin{tabular}[width=\textwidth]{c|c|c|c|c}
\textbf{Metadata} & \textbf{Van der Pol} & \textbf{Hod.-Hux.}  & \textbf{Nest. LCO }  &  \textbf{Lorenz-96 (F=2.75)}  \\
\hline
\# Traj.   & 400 & 500 & 600 & 400  \\
Traj. length (s)  & 45  &  44 & 50 & 25\\
$\Delta$t  (s) & 0.05 & 0.04 & 0.05 & 0.05  \\
$\tau_d$ (s) & 0.95 & 2.0 & 0.95 & 0.95 \\
GPU & RTX 3090 & RTX 3090  & RTX 3090 & RTX 3090 \\
Learn. rate (min.) & $3 \cdot 10^{-4}$ & $3 \cdot 10^{-4}$ & $3 \cdot 10^{-4}$  & $3\cdot10^{-4}$\\
Learn rate (max.) & $3\cdot 10^{-3}$ & $3 \cdot 10^{-3}$ & $3 \cdot 10^{-3}$   & $1\cdot10^{-3}$\\
Batch size & 1000 & 1000 & 1000 & 128 \\
\# Pretraining epochs & 5 & 5 & 5 & 5 \\
\# Main epochs & 120 & 105 & 120 & 100 \\
Train./Val. $T$ (s) & 20/24 & 18/22 & 20/24 & 11/13.5 \\
\end{tabular}
\caption{Table outlining the metadata for the dimensionality experiments for the Van der Pol oscillator, the Hodgkin-Huxley model, the nested limit-cycle oscillator and the Lorenz 96 system with F=2.75.}\label{tab: Studies table 1}
\end{table}

\begin{table}[h]
\centering
\begin{tabular}[width=\textwidth]{c|c|c|c}
\textbf{Metadata} & \textbf{Duffing} & \textbf{MMSD}  & \textbf{Pend.}    \\
\hline
\# Traj.   & 600 & 800 & 800  \\
Traj. length (s)  & 50  &  40 & 40  \\
$\Delta$t  (s) & 0.05 & 0.04 & 0.05   \\
$\tau_d$ (s) & 1.0 & 1.6 & 1.05   \\
GPU & RTX 3090 & RTX A6000 & RTX A6000 \\
Learn. rate (min)& $3\cdot 10^{-4}$ & $3\cdot 10^{-4}$ & $2\cdot10^{-4}$ \\
Learn. rate (max)& $5\cdot10^{-3}$ & $5\cdot10^{-3}$ & $3\cdot 10^{-3}$ \\
Batch size & 1000 & 1000 & 2400 \\
\# Pretraining epochs & 5 & 5 & 5 \\
\# Main epochs & 100 & 100 & 130 \\
Train./Val. $T$ (s) & 20/25 & 16/20 & 20/22.5 \\
\end{tabular}
\caption{Table outlining the metadata for the dimensionality experiments for the Duffing equation, the magnetic mass-spring-damper (MMSD), and the single pendulum.}\label{tab: Studies table 2}
\end{table}

\begin{table}[h]
\centering
\begin{tabular}[width=\textwidth]{c|c|c|c}
\textbf{Metadata}  & \textbf{Exp. Mag. Pend.}   &  \textbf{Exp. Doub. Pend.} & \textbf{Lorenz-96 (F=8.0)}   \\
\hline
\# Traj.   & 900 & 795 & 1250  \\
Traj. length (s)  & 10  &  26 & 20  \\
$\Delta$t  (s) & 0.02 & 0.02 & 0.02 \\
$\tau_d$ (s) & 0.62 & 1.98 & 0.80  \\
GPU & RTX A6000 & RTX A6000 & RTX A6000  \\
Learn. rate (min)& $3\cdot 10^{-4}$ & $1\cdot 10^{-3}$ & N/A  \\
Learn. rate (max)& $3\cdot10^{-3}$ & $3\cdot10^{-3}$ & $2.19\times 10^{-3}$ \\
Batch size & 500 & 1200 & 1000 \\
\# Pretraining epochs & 20 & 5 & 8 \\
\# Main epochs & 200 & 100 & 130  \\
Train./Val. $T$ (s) & 6/6.6 & 3/3.6 & 0.8/0.8 \\
\end{tabular}
\caption{Table outlining the metadata for the model training and dimensionality experiments for the experimental magnetic pendulum, the experimental double pendulum, and the chaotic Lorenz-96 system}\label{tab: Studies table 3}
\end{table}

\begin{table}[h]
\centering
\begin{tabular}[width=\textwidth]{c|c|c|c|c}
\textbf{Metadata} & \textbf{Annealing} & \textbf{Ablation} & \textbf{MI} & \textbf{MI}  \\
\hline
System & Duffing & MMSD & Pend. & Doub. pend.  \\
\# Train./Test. traj.   & 350/150 & 700/300 & 560/240 & 1050/450    \\
Traj. length (s)  & 50  &  40 & 40 & 40 \\
$\Delta$t  (s) & 0.05 & 0.04 & 0.05 & 0.04  \\
$\tau_d$ (s) & 1.0 & 1.6 & N/A & N/A \\
Added noise $\sigma$ & 0  & 0 & 0.01 & 0.03 \\
Latent dim. & 5 & 5 & 3 & 60\\
Dropout rate & 0.001 & 0.001 & 0.001 & 0.001  \\
GPU & RTX 3090 & RTX A6000 & RTX A6000 & RTX A6000 \\
Learn. rate (min) & $4\cdot 10^{-3}$ & $4\cdot 10^{-3}$ & $2\cdot 10^{-3}$  & $1\cdot 10^{-3}$\\
Learn. rate (max) & $4\cdot 10^{-3}$ & $4\cdot 10^{-3}$ & $2\cdot 10^{-3}$  & $1\cdot 10^{-3}$\\
Batch size & 1000 & 1000 & 2400 & 5000 \\
\# Pretraining epochs & 5 & 5 & 5 & 8 \\
\# Main epochs & 100 & 100 & 100 & 130 \\
Train./Test. $T$ (s) & 20/30 & 16/20 & 15/20 & 2/2 \\
\# Trials (each strategy) & 25 & 25 & 25 & 25\\
\end{tabular}
\caption{Table outlining the metadata for the annealing, ablation, and mutual information experiments.}\label{tab: Studies table 4: annealing ablation MI}
\end{table}

\begin{table}[h]
\centering
\begin{tabular}{c|ccc}
  & \textbf{No Anneal.} & \textbf{Std. Anneal.} & \textbf{Cyc. Anneal.}  \\
 \textbf{Annealing Parameter Bounds} & ($\alpha_{min}$, $\alpha_{max}$) & ($R_{min}$, $R_{max}$) & ($M_{min}$, $M_{max}$)  \\
\hline
$\mathcal{L}_{x_0}$   & (1.0, 1.0) & (1.0, 1.0) & (1.0, 1.0) \\
$\mathcal{L}_{\dot{\psi}}$   & (0.01, 1.0) & (0.2, 0.8) & (2, 8) \\
$\mathcal{L}_{\psi_t}$   & (0.01, 1.0)   & (0.2, 0.8) & (2, 8) \\
$\mathcal{L}_{x_t}$  & (0.01, 1.0)  & (0.2, 0.8) &  (2, 8) \\
$\mathcal{L}_{\mu}$  & (0.01, 1.0)  & (0.2, 0.8) & (2, 8) \\
$\gamma$ & (0.9, 1.0)  & (0.5, 0.9) & (2, 5) \\
\end{tabular}
\caption{This table gives bounds for the annealing hyperparameters used for all of the models trained with systematic annealing.}\label{tab: annealing table}
\end{table}

\begin{table}[h]
\centering
\begin{tabular}{c|c|c}
\textbf{Parameter} & \textbf{Value} & \textbf{Description} \\
\hline
$g$ & 9.81 & Gravitational acceleration (m/s$^2$) \\
$l$ & 1.0 & Length of the pendulum (m) \\
$\beta$ & 0.2 & Damping coefficient \\
\end{tabular}
\caption{The values for parameters in Eq. \ref{eq: pend eom}, the pendulum equations of motion, used in numerical simulation.}
\label{tab: single pend tab}
\end{table}

\begin{table}[h]
\centering
\begin{tabular}{c|c|c}
\textbf{Parameter} & \textbf{Value} & \textbf{Description} \\
\hline
$g$ & 9.81 & Gravitational acceleration (m/s$^2$) \\
$l_1$ & 1.0 & Length of the first pendulum (m) \\
$l_2$ & 1.0 & Length of the second pendulum (m) \\
$m_1$ & 1.0 & Mass of the first pendulum (kg) \\
$m_2$ & 2.0 & Mass of the second pendulum (kg) \\
$\beta$ & 0.1 & Damping coefficient \\
\end{tabular}
\caption{The values for parameters in Eq. \ref{eq: double pend eom} used for numerical simulation.}
\label{tab: double pend table}
\end{table}

\begin{table}[h]
\centering
\begin{tabular}{c|c|c}
\textbf{Parameter} & \textbf{Value} & \textbf{Description} \\
\hline
$m$ & 1.0 & Mass (kg) \\
$c$ & 0.5 & Damping coefficient (Ns/m)\\
$k$ & 10.0 & Spring constant (N/m) \\
$\alpha$ & 100.0 & Magnetic force coefficient \\
$h$ & 1.5 & Vertical displacement between magnets (m) \\
$b$ & 1.3 & Initial horizontal offset (m) \\
\end{tabular}
\caption{The values for parameters used in numerical simulation of Eq. \ref{eq: mmsd eom}.}
\label{tab: mmsd tab}
\end{table}

\end{document}